\documentclass[11pt]{article}
\usepackage[margin=1in]{geometry}

\usepackage{macros}
\bluehyperref

\usepackage[style=alphabetic,backend=bibtex,maxnames=12,maxbibnames=10,maxcitenames=10,maxalphanames=10,giveninits=true,doi=false,url=true]{biblatex}
\newcommand*{\citet}[1]{\AtNextCite{\AtEachCitekey{\defcounter{maxnames}{2}}} \textcite{#1}}

\newcommand*{\citep}[1]{\cite{#1}}
\newcommand{\citeyearpar}[1]{\cite{#1}}
\bibliography{bib, vf-allrefs-local, compression-refs}
\let\citealp\citep

\usepackage{cleveref}

\usepackage{overpic}

\usepackage[noend]{algorithmic}
\usepackage{algorithm}

\newif\ificml

\newtheorem{theorem}{Theorem}
\newtheorem{lemma}{Lemma}[section]
\newtheorem{proposition}{Proposition}
\newtheorem{definition}{Definition}[section]

\newtheorem*{lemma*}{Lemma}

\renewcommand{\R}{\mathcal{R}} 
\renewcommand{\r}{\mathbb{R}} 
\newcommand{\privunit}{\text{PrivUnit}} 
\newcommand{\pu}{\privunit} 
\newcommand{\pug}{\text{PrivUnitG}} 
\newcommand{\err}{\mathsf{Err}} 
\newcommand{\pdf}{f} 

\newcommand{\AdA}{\A^+} 

\newcommand{\eqd}{\stackrel{d}{=}} 

\title{Optimal Algorithms for Mean Estimation under Local Differential Privacy}
\author{%
    Hilal Asi\thanks{Stanford University, part of this work performed while interning at Apple; \texttt{asi@stanford.edu}.}
    \and Vitaly Feldman\thanks{Apple; \texttt{vitaly.edu@gmail.com}.}
    \and Kunal Talwar\thanks{Apple; \texttt{kunal@kunaltalwar.org}.}
    }

\begin{document}

\maketitle

\begin{abstract}
    We study the problem of mean estimation of $\ell_2$-bounded vectors under the constraint of local differential privacy. While the literature has a variety of algorithms that achieve the asymptotically optimal rates for this problem, the performance of these algorithms in practice can vary significantly due to varying (and often large) hidden constants. In this work, we investigate the question of designing the protocol with the smallest variance. We show that PrivUnit~\cite{BhowmickDuFrKaRo18} with optimized parameters achieves the optimal variance among a large family of locally private randomizers. To prove this result, we establish some properties of local randomizers, and use symmetrization arguments that allow us to write the optimal randomizer as the optimizer of a certain linear program. These structural results, which should extend to other problems, then allow us to show that the optimal randomizer belongs to the PrivUnit family.
    
    We also develop a new variant of PrivUnit based on the Gaussian distribution which is more amenable to mathematical analysis and enjoys the same optimality guarantees. This allows us to establish several useful properties on the exact constants of the optimal error as well as to numerically estimate these constants.
    
\end{abstract}

\section{Introduction}

Mean estimation is one of the most fundamental problems in machine learning and is the building block of a countless number of algorithms and applications including stochastic optimization~\cite{Duchi18}, federated learning~\cite{BonawitzIvKrMaMcPaRaSeSe17} and others. However, it is now evident that standard algorithms for this task may leak sensitive information about users' data and compromise their privacy. This had led to the development of numerous algorithms for estimating the mean while preserving the privacy of users. The most common models for privacy are either the central model where there exists a trusted curator or the local model where such trusted curator does not exist.

In this work, we study the problem of mean estimation in the local privacy model. More specifically, we have $n$ users each with a vector $v_i$ in the Euclidean unit ball in $\r^d$. Each user will use a randomizer $\R: \r^d \to \domain$ to privatize their data where $\R$ must satisfy $\diffp$-differential privacy, namely, for any $v_1$ and $v_2$, $P(\R(v_1)=u)/P(\R(v_2)=u) \le e^\diffp$. Then, we run an aggregation method $\A : \domain^n \to \r^d$ such that $\A(\R(v_1),\dots,\R(v_n))$  provides an estimate of $\frac{1}{n} \sum_{i=1}^n v_i$.
Our goal in this work is to characterize the optimal protocol (pair of randomizer $\R$ and aggregation method $\A$) for this problem and study the resulting optimal error.

Due to its importance and many applications, the problem of private mean estimation in the local model has been studied by numerous papers~\cite{BhowmickDuFrKaRo18,FeldmanTa21,ChenKaOz20}. As a result, a clear understanding of the asymptotic optimal rates has emerged, showing that the optimal squared error is proportional to $\Theta(\frac{d}{n\min(\eps, \eps^2)})$: \citet{DuchiJoWa18, BhowmickDuFrKaRo18} developed algorithms that obtain this rate and ~\cite{DuchiRo19} proved corresponding lower bounds. Subsequent papers~\cite{FeldmanTa21,ChenKaOz20} have developed several other algorithms that achieve the same rates.

However, these optimality results do not give a clear characterization of which algorithm will enjoy better performance in practice. Constant factors here matter more than they do in run time or memory, as $\eps$ is typically limited by privacy constraints, and increasing the sample size by collecting data for more individuals is often infeasible or expensive. The question of finding the randomizer with the smallest error is therefore of great interest. 

\subsection{Our contributions}
Motivated by these limitations, we investigate strict optimality for the problem of mean estimation with local privacy. We study the family of non-interactive and unbiased protocols, that is, a protocol is a pair of local private randomizer $\R : \sphere^{d-1} \to \domain$ and an aggregation method $\A : \domain^n \to \r^d$ where the protocol outputs $\A(\R(v_1),\dots,\R(v_n))$ such that $\E[\A(\R(v_1),\dots,\R(v_n))] = \frac{1}{n} \sum_{i=1}^n v_i$.  We measure the error of a private protocol in terms of its (worst case) mean squared error 
\begin{equation*}
    \err_n(\A,\R) =  \sup_{v_1,\dots,v_n \in \sphere^{d-1}} \E \left[ \ltwo{\A(\R(v_1),\dots,\R(v_n)) - \frac{1}{n} \sum_{i=1}^n v_i}^2 \right].
\end{equation*}
We obtain the following results. 

First, we show that $\pu$ of~\citet{BhowmickDuFrKaRo18} with optimized parameters is optimal amongst a large family of protocols. Our strategy for proving optimality consists of two main steps: first, we show that for non-interactive protocols, additive aggregation with a certain randomizer attains the optimal error. Then, for protocols with additive aggregation, we show that $\pu$ obtains the optimal error. Our proof builds on establishing several new properties of the optimal local randomizer, which allow us to express the problem of designing the optimal randomizer as a linear program. This in turn helps characterize the structure of optimal randomizers and allows us to show that there is an optimal randomizer which is an instance of $\pu$.

Finding the exact constants in the error of $\pu$ is mathematically challenging. Our second contribution is to develop a new algorithm $\pug$ that builds on the Gaussian distribution and attains the same error as $\pu$ up to a $(1+o(1))$ multiplicative factor as $d \to \infty$. In contrast to $\pu$, we show that the optimal parameters of $\pug$ are independent of the dimension, hence enabling efficient calculation of the constants for high dimensional settings. Moreover, $\pug$ is amenable to mathematical analysis which yields several properties on the constants of the optimal error.

\subsection{Related work}

Local privacy is perhaps one of the oldest forms of privacy and dates back to~\citet{Warner65} who used it to encourage truthfulness in surveys. This definition resurfaced again in the contex of modern data analysis by~\citet{EvfimievskiGeSr03} and was related to differential privacy in the seminal work of ~\citet{DworkMNS:06}. Local privacy has attracted a lot of interest, both in the academic community~\cite{BeimelNiOm08,DuchiJoWa18,BhowmickDuFrKaRo18}, and in industry where it has been deployed in several industrial applications~\cite{ErlingssonPiKo14, ApplePrivacy17}. Recent work in the Shuffle model of privacy~\citep{Bittau17, Cheu:2019, ErlingssonFMRTT19, BalleBGN19a, FeldmanMcTa20} has led to increased interest in the local model with moderate values of the local privacy parameter, as they can translate to small values of central $\eps$ under shuffling.

The problem of locally private mean estimation has received a great deal of attention in the past decade~\cite{DuchiJoWa18,BhowmickDuFrKaRo18,DuchiRo19,ErlingssonFeMiRaSoTaTh20,AgarwalSYKM18, girgis2020shuffled, ChenKaOz20, GKMM19,FeldmanTa21}. \citet{DuchiJoWa18} developed asymptotically optimal procedures for estimating the mean when $\diffp \le 1$, achieving expected squared error $O(\frac{d}{n \diffp^2})$. \citet{BhowmickDuFrKaRo18} proposed a new algorithm that is optimal for $\diffp \ge 1$ as well, achieving error $O(\frac{d}{n\min(\eps, \eps^2)})$. These rates are optimal as~\citet{DuchiRo19} show tight lower bounds which hold for interactive protocols. There has been more work on locally private mean estimation that studies the problem with additional constraints such as communications cost~\cite{ErlingssonFeMiRaSoTaTh20,FeldmanTa21,ChenKaOz20}.

\citet{YeBa17,YeBa18} study (non-interactive) locally private estimation problems with discrete domains and design algorithms that achieve optimal rates. These optimality results are not restricted to the family of unbiased private mechanisms. However, in contrast to our work, these results are only asymptotic hence their upper and lower bounds are matching only as the number of samples goes to infinity. 

While there are several results in differential privacy that establish asymptotically matching lower and upper bounds for various problems of interest, strict optimality results are few. While there are some results known for the one-dimensional problem~\cite{GhoshRS09, GupteR10}, some of which extend to a large class of utility functions, such universal mechanisms are known not to exist for multidimensional problems~\cite{BrennerN2010}. \cite{GengKOV15, KOV16jmlr} show that for certain loss functions, one can phrase the problem of designing optimal local randomizers as linear programs, whose size is exponential in the size of the input domain.

\section{Problem setting and preliminaries}

We begin in this section by defining local differential privacy. To this end, we say that two probability distributions $P$ and $Q$ are \ed-close if for every event $E$ 
\begin{equation*}
    e^{-\diffp} (P(E) - \delta) 
        \le Q(E) 
        \le e^{\diffp} P(E) + \delta. 
\end{equation*}
We say two random variables are \ed-close if their distributions are \ed-close.

We can now define local DP randomizers. 
\begin{definition}
    A randomized algorithm $\R : X \to Y$ is (replacement) \ed-DP local randomizer if for all $x,x' \in X$, $\R(x)$ and $\R(x')$ are \ed-close.
\end{definition}

In this work, we will primarily be interested in {\em pure} DP randomizers, i.e. those which satisfy $(\eps, 0)$-DP. We abbreviate this as $\eps$-DP. In the setting of local randomizers, the difference between \ed-DP and pure DP is not significant; indeed any \ed-DP local randomizer can be converted~\cite{FeldmanMcTa20, CheuU21} to one that satisfies $\eps$-DP while changing the distributions by a statistical distance of at most $O(\delta)$.

The main problem we study in this work is locally private mean estimation. Here, we have $n$ unit vectors $v_1,\dots,v_n \in \r^d$, i.e $v_i \in \sphere^{d-1}$. The goal is to design (locally) private protocols that estimate the mean $\frac{1}{n} \sum_{i=1}^n v_i$. We focus on the setting of non-interactive private protocols: such a protocol consists of a pair of private local randomizer $\R: \sphere^{d-1} \to \domain$ and aggregation method $\A: \domain^n \to \r^d$ where the final output is $\A(\R(v_1),\dots,\R(v_n))$. We require that the output is unbiased, that is, $\E[\A(\R(v_1),\dots,\R(v_n))] = \frac{1}{n} \sum_{i=1}^n v_i$, and wish to find private protocols that minimize the variance
\begin{equation*}
    \err_n(\A,\R) =  \sup_{v_1,\dots,v_n \in \sphere^{d-1}} \E \left[ \ltwo{\A(\R(v_1),\dots,\R(v_n)) - \frac{1}{n} \sum_{i=1}^n v_i}^2 \right].
\end{equation*}

Note that in the above formulation, the randomizer $\R$ can have arbitrary domains (not necessarily $\r^d$), and the aggregation method can be arbitrary as well. However, one important special family of private protocols, which we term \emph{canonical private protocols}, are protocols where the local randomizer $\R : \sphere_2^{d-1} \to \r^d$ has outputs in $\r^d$ and the aggregation method is the simple additive aggregation $\AdA(z_1,\dots,z_n) = \frac{1}{n} \sum_{i=1}^n z_i$. In addition to being a natural family of protocols, canonical protocols are $(i)$ simple and easy to implement, and $(ii)$ achieve the smallest possible variance amongst the family of all possible unbiased private protocols, as we show in the subsequent sections.

\paragraph{Notation} 
We let $\sphere^{d-1} = \{u \in \r^d : \ltwo{u} = 1 \}$ denote the unit sphere, and $R \cdot \sphere^{d-1}$ denote the sphere of radius $R>0$. Whenever clear from context, we use the shorter notation $\sphere$.
Given a random variable $V$, we let $\pdf_{V}$ denote the probability density function of $V$. For a randomizer $\R$ and input $v$, $\pdf_{\R(v)}$ denotes the probability density function of the random variable $\R(v)$. For a Gaussian random variable $V \sim \normal(0,\sigma^2)$ with $\sigma>0$, we let $\phi_\sigma: \r \to \r$ denote the probability density function of $V$ and $\Phi_\sigma: \r \to [0,1]$ denote its cumulative distribution function. For ease of notation, we write $\phi$ and $\Phi$ when $\sigma=1$. Given two random variables $V$ and $U$, we say that $V \eqd U$ if $V$ and $U$ has the same distribution, that is, $\pdf_V = \pdf_U$. Finally, we let $e_i \in \r^d$ denote the standard basis vectors and $O(d) = \{U \in \r^{d \times d}: U U^T = I \}$ denote the subspace of orthonormal matrices of dimension $d$.

\section{Optimality of PrivUnit}
\label{sec:opt}
In this section, we prove our main optimality results showing that $\pu$ with additive aggregation achieves the optimal error among the family of unbiased locally private procedures. More precisely, we show that for any $\diffp$-DP local randomizer $\R: \r^d \to \domain$ and any aggregation method $\A : \domain^n \to \r^d$ that is unbiased, 
\begin{equation*}
    \err_n(\AdA,\pu) \le 
    \err_n(\A,\R).
\end{equation*}
We begin in~\Cref{sec:pu-def} by introducing the algorithm $\pu$ and stating its optimality guarantees in~\Cref{sec:pu-opt}. To prove the optimality result, we begin in~\Cref{sec:canonical-opt} by showing that there exists a canonical private protocol that achieves the optimal error, then in~\Cref{sec:pu-opt} we show that $\pu$ is the optimal local randomizer in the family of canonical protocols.

\subsection{$\pu$}
\label{sec:pu-def}
We begin by introducing $\pu$ which was developed by~\citet{BhowmickDuFrKaRo18}. Given an input vector $v \in \sphere^{d-1}$ and letting $W \sim \uniform(\sphere^{d-1})$, $\pu(p,\gamma)$ has the following distribution (up to normalization) 
\begin{equation*}
\pu(p,\gamma) \sim 
        \begin{cases}
            & W \mid \<W,v\> \ge \gamma  \quad  \text{with prob. } p \\
            & W \mid \<W,v\> < \gamma \quad  \text{with prob. } 1-p
        \end{cases}
\end{equation*}
A normalization factor is needed to obtain the correct expectation. We provide full details in~\Cref{alg:pu2}. 

The following theorem states the privacy guarantees of $\pu$. Theorem 1 in~\cite{BhowmickDuFrKaRo18} provides privacy guarantees based on several mathematical approximations which may not be tight. For our optimality results, we require the following exact privacy guarantee of $\pu$.
\ificml
\begin{theorem}~\citealp[Theorem 1]{BhowmickDuFrKaRo18}
\label{thm:pu-privacy}
    Let $q = P(W_1 \le \gamma)$ where $W \sim \uniform(\sphere^{d-1})$. If $\frac{p}{1-p}\frac{q}{1-q} \le e^\diffp$ then $\pu(p,\gamma)$ is an $\diffp$-DP local randomizer.
\end{theorem}
\else
\begin{theorem}~\cite[Theorem 1]{BhowmickDuFrKaRo18}
\label{thm:pu-privacy}
    Let $q = P(U_1 \le \gamma)$ where $W \sim \uniform(\sphere^{d-1})$. If $\frac{p}{1-p}\frac{q}{1-q} \le e^\diffp$ then $\pu(p,\gamma)$ is an $\diffp$-DP local randomizer.
\end{theorem}
\fi

Throughout the paper, we will sometimes use the equivalent notation $\pu(p,q)$ which describes running $\pu(p,\gamma)$ with $q = P(W_1 \le \gamma)$ as in~\Cref{thm:pu-privacy}.

\ificml
\else
\begin{algorithm}
	\caption{$\pu(p,\gamma)$}
	\label{alg:pu2}
	\begin{algorithmic}[1]
		\REQUIRE $v \in \sphere^{d-1}$, $\gamma \in [0,1]$, $p \in [0,1]$. $B(\cdot; \cdot, \cdot)$ below is the incomplete Beta function $B(x;a,b) = \int_0^x t^{a-1}(1-t)^{b-1} \textrm{d}t$ and $B(a,b) = B(1; a, b)$.
		\STATE Draw $z \sim \mathsf{Ber}(p)$
		\IF{$z=1$}
		    \STATE Draw $V \sim \uniform \{u \in \sphere^{d-1}: \< u,v\> \ge \gamma \}$
		\ELSE
		     \STATE Draw $V \sim \uniform \{u \in \sphere^{d-1}: \< u,v\> < \gamma \}$
		\ENDIF
		\STATE Set $\alpha = \frac{d-1}{2}$ and $\tau = \frac{1 + \gamma}{2}$
		\STATE Calculate normalization constant
		\begin{equation*}
		    m = \frac{(1-\gamma^2)^\alpha}{2^{d-2} (d-1)} \left( \frac{p}{B(\alpha,\alpha) - B(\tau; \alpha,\alpha)} + \frac{1-p}{B(\tau; \alpha,\alpha)}   \right)
		\end{equation*}
        \STATE Return $\frac{1}{m} \cdot V$
	\end{algorithmic}
\end{algorithm}
\fi

\subsection{Optimality}
\label{sec:pu-opt}
Asymptotic optimality of $\pu$ has already been established by prior work.
\citet{BhowmickDuFrKaRo18} show that the error of $\pu$ is upper bounded by $O(\frac{d}{n \min(\diffp,\diffp^2)})$ for certain parameters. Moreover, \citet{DuchiRo19} show a lower bound of $\Omega(\frac{d}{n \min(\diffp,\diffp^2)})$, implying that $\pu$ is asymptotically optimal.

In this section, we prove that additive aggregation applied with $\pu$ with the best choice of parameters $p,\gamma$ is truly optimal, that is, it outperforms any unbiased private algorithm. 
The following theorem states our optimality result for $\pu$.
\begin{theorem}
\label{thm:opt-general}
    Let $\R : \sphere^{d-1} \to \domain $ be an $\diffp$-DP local randomizer, and $\A : \domain^n \to \r^d$ be an aggregation procedure such that $\E[\A(\R(v_1),\dots,\R(v_n))] = \frac{1}{n} \sum_{i=1}^n v_i$ for all $v_1,\dots,v_n \in \sphere^{d-1}$. Then there is $p^\star \in [0,1]$ and $\gamma^\star \in [0,1]$ such that $\pu(p^\star,\gamma^\star)$ is $\diffp$-DP local randomizer and
    \begin{equation*}
        \err(\AdA,\pu(p^\star,\gamma^\star)) \le \err(\A,\R).
    \end{equation*}
\end{theorem}
The proof of~\Cref{thm:opt-general} will proceed in two steps: first, in~\Cref{sec:canonical-opt} (\Cref{thm:opt-agg}), we show that there exists an optimal private procedure that is canonical, then in~\Cref{sec:pu-opt} (\Cref{thm:opt-repl}) we prove that $\pu$ is the optimal randomizer in this family. \Cref{thm:opt-general} is a direct corollary of these two propositions.


\subsection{Optimality of canonical protocols}
\label{sec:canonical-opt}
In this section, we show that there exists a canonical private protocol that achieves the optimal error. In particular, we have the following result.

\begin{proposition}
\label{thm:opt-agg}
    Let $(\R,\A)$ be such that $\R: \sphere^{d-1} \to \domain$ is $\diffp$-DP local randomizer and $\E[\A(\R(v_1),\dots,\R(v_n))] = \frac{1}{n} \sum_{i=1}^n v_i$ for all $v_1,\dots,v_n \in \sphere^{d-1}$. Then there a canonical randomizer $\R': \sphere^{d-1} \to \r^d$ that is $\diffp$-DP local randomizer and
    \begin{equation*}
        \err_n(\A,\R) \ge \err_n(\AdA,\R').
    \end{equation*}
\end{proposition}

To prove~\Cref{thm:opt-agg}, we begin with the following lemma.
\begin{lemma}
\label{lemma:var-decomp}
    Let $(\R,\A)$ satisfy the conditions of~\Cref{thm:opt-agg}.
    Let $P$ be a probability distribution over $\sphere^{d-1}$ such that $\E_{v \sim P}[v] = 0$.
    There is $\hat \R_i : \sphere^{d-1} \to \r^d$ for $i \in [n]$ such that $\hat \R_i$ is $\diffp$-DP local randomizer, $\E[\hat \R_i(v)] = v$ for all $v \in \sphere^{d-1}$, and
    \begin{equation*}
     \E_{v_1,\dots,v_n \sim P} \left[ \ltwo{n \A(\R(v_1),\dots,\R(v_n)) -  \sum_{i=1}^n v_i}^2 \right]
        \ge \sum_{i=1}^n \E_{v_i \sim P} \left[ \ltwo{\hat \R_i(v_i) - v_i }^2 \right].  
    \end{equation*}
\end{lemma}

\newcommand{\Pu}{P_{\mathsf{unif}}}
Before proving~\Cref{lemma:var-decomp}, we now complete the proof the~\Cref{thm:opt-agg}.
\begin{proof}(\Cref{thm:opt-agg})
    Let $\Pu$ be the uniform distribution over the sphere $\sphere^{d-1}$. First, note that 
    \begin{align*}
     n^2 \err_n(\A,\R) 
        & = \sup_{v_1,\dots,v_n \in \sphere^{d-1}}\E \left[ \ltwo{n \A(\R(v_1),\dots,\R(v_n)) -  \sum_{i=1}^n v_i}^2 \right] \\
        & \ge \E_{v_1,\dots,v_n \sim \Pu} \left[ \ltwo{n \A(\R(v_1),\dots,\R(v_n)) -  \sum_{i=1}^n v_i}^2 \right] \\
        &  \ge \sum_{i=1}^n \E_{v_i \sim \Pu} \left[ \ltwo{\hat \R_i(v_i) - v_i }^2 \right].  
    \end{align*}
    Now we define $\R'_i$ as follows. 
    First, sample a random rotation matrix $U \in \r^{d \times d}$ where $U^T U = I$, then set
    \begin{equation*}
        \R'_i(v) = U^T \hat\R_i(Uv).
    \end{equation*}
    Note that $\R'$ is $\diffp$-DP local randomizer, $\E[\R'_i(v)] = v$, and for all $v \in \sphere^{d-1}$
    \begin{align*}
    \E\left[\ltwo{\R'_i(v) - v}^2 \right] 
        & = \E_{U} \left[\ltwo{U^T\hat\R_i(Uv) - v}^2\right] \\
        & = \E_{U} \left[\ltwo{\hat\R_i(Uv) - U v}^2\right] \\
        & = \E_{v \sim \Pu } \left[\ltwo{\hat\R_i(v) - v}^2\right].
    \end{align*}
    Overall, we have
    \begin{align*}
     n^2 \err_n(\A,\R) 
        &  \ge \sum_{i=1}^n \E_{v_i \sim \Pu} \left[ \ltwo{\hat \R_i(v_i) - v_i }^2 \right] \\
        & = \sum_{i=1}^n \sup_{v \in \sphere^{d-1}} \E \left[\ltwo{\R'_i(v) - v}^2 \right] \\
        & = \sum_{i=1}^n \err_1(\AdA,\R'_i) \\
        & \ge n \err_1(\AdA,\R'_{i^\star}) \\
        & = n^2 \err_n(\AdA,\R_{i^\star}),
    \end{align*}
    where $i^\star \in [n]$ minimizes $\err_1(\AdA,\R'_i)$.
    The claim follows.
\end{proof}

Now we prove~\Cref{lemma:var-decomp}. 
\begin{proof}(\Cref{lemma:var-decomp})
    We define $\hat \R_i$ to be
    \begin{equation*}
        \hat \R_i(v_i) = \E_{v_j \sim P, j \neq i} [n \A(\R(v_1),\dots,\R(v_n))] .
    \end{equation*}
    Note that $\hat \R_i$ is $\diffp$-DP local randomizer as it requires a single application of $\R(v_i)$. Moreover, $\E[\hat R_i(v)] = v$ for all $v \in \sphere^{d-1}$. We define 
    \begin{equation*}
        \hat \R_{\le i}(v_1,\dots,v_i) = \E_{v_j \sim P, j > i} \left[n\A(\R(v_1),\dots,\R(v_n)) - \sum_{j=1}^i v_j \right],
    \end{equation*}
    and $\hat \R_0 = 0$.
    We now have
    \begin{align*}
     & \E_{v_1,\dots,v_n \sim P} \left[ \ltwo{n \A(\R(v_1),\dots,\R(v_n)) -  \sum_{i=1}^n v_i}^2 \right]  \\
        & \quad = \E_{v_1,\dots,v_n \sim P} \left[ \ltwo{ \hat \R_{\le n}(v_1,\dots,v_n)   }^2 \right] \\
        &  \quad = \E_{v_1,\dots,v_n \sim P} \left[ \ltwo{ \hat \R_{\le n}(v_1,\dots,v_n) - \hat \R_{\le n-1}(v_1,\dots,v_{n-1}) + \hat \R_{\le n-1}(v_1,\dots,v_{n-1})    }^2 \right] \\
        & \quad  \stackrel{(i)}{=} \E_{v_1,\dots,v_n \sim P} \left[ \ltwo{ \hat \R_{\le n}(v_1,\dots,v_n) - \hat \R_{\le n-1}(v_1,\dots,v_{n-1}) }^2 \right] +  \E_{v_1,\dots,v_{n-1} \sim P} \left[ \ltwo{\hat \R_{\le n-1}(v_1,\dots,v_{n-1})    }^2 \right] \\
        & \quad  \stackrel{(ii)}{=} \sum_{i=1}^n \E_{v_1,\dots,v_i \sim P} \left[ \ltwo{ \hat \R_{\le i}(v_1,\dots,v_i) - \hat \R_{\le i-1}(v_1,\dots,v_{i-1}) }^2 \right] \\
        & \quad  \stackrel{(iii)}{\ge} \sum_{i=1}^n \E_{v_i \sim P} \left[ \ltwo{ E_{v_1,\dots,v_{i-1} \sim P} \left[ \hat \R_{\le i}(v_1,\dots,v_i) - \hat \R_{\le i-1}(v_1,\dots,v_{i-1})\right] }^2 \right] \\
        & \quad  \stackrel{(iv)}{=} \sum_{i=1}^n \E_{v_i \sim P} \left[ \ltwo{\hat \R_i(v_i) - v_i}^2 \right] \\
    \end{align*}
    where $(i)$ follows since $E_{v_n \sim P} [\R_{\le n}(v_1,\dots,v_n)] = \hat \R_{\le n-1}(v_1,\dots,v_{n-1})$, $(ii)$ follows by induction, $(iii)$ follows from Jensen's inequality, and $(iv)$ follows since $E_{v_1,\dots,v_{i-1} \sim P} [ \hat \R_{\le i}(v_1,\dots,v_i)] = \hat \R_i(v_i) - v_i$ and $E_{v_1,\dots,v_{i-1} \sim P} [ \hat \R_{\le i-1}(v_1,\dots,v_{i-1})] = 0$.

\end{proof}

\subsection{Optimality of $\pu$ among canonical randomizers}
\label{sec:pu-opt-can}
In this section, we show that $\pu$ achieves the optimal error in the family of canonical randomizers. To this end, first note that for additive aggregation $\AdA$, we have $\err_n(\AdA,\R) = \err_1(\AdA,\R)/n$. Denoting $\err(\R) = \err_1(\AdA,\R)$ for canonical randomizers, we have the following optimality result.

\begin{proposition}
\label{thm:opt-repl}
    Let $\R : \sphere^{d-1} \to \r^d $ be an $\diffp$-DP local randomizer such that $\E[\R(v)] = v$ for all $v \in \sphere^{d-1}$. Then there is $p^\star \in [0,1]$ and $\gamma^\star \in [0,1]$ such that $\pu(p^\star,\gamma^\star)$ is $\diffp$-DP local randomizer and
    \begin{equation*}
        \err(\pu(p^\star,\gamma^\star)) \le \err(\R).
    \end{equation*}
\end{proposition}


The proof of~\Cref{thm:opt-repl} builds on a sequence of lemmas, each of which allows to simplify the structure of an optimal algorithm. 
We begin with the following lemma which show that there exists an optimal algorithm which is invariant to rotations.
\begin{lemma}[Rotation-Invariance Lemma]
\label{lemma:symmetry}
    Let $\R : \sphere^{d-1} \to \r^d $ be an $\diffp$-DP local randomizer such that $\E[\R(v)] = v$ for all $v \in \sphere^{d-1}$. There exists an $\diffp$-DP local randomizer $\R'$ such that
    \begin{enumerate}
        \item $\E[\R'(v)] = v$ for all $v \in \sphere^{d-1}$
        \item $\err(\R') \le \err(\R)$
        \item $\err(\R') = \err(\R',v)$ for all $v \in \sphere^{d-1}$
        \item For any $v, v_0 \in \sphere^{d-1}$, there is an orthonormal matrix $V \in \r^{d \times d}$ such that $\R'(v) \eqd V \R'(v_0)$.
        \item $\frac{\pdf_{\R'(v)}(u_1)}{\pdf_{\R'(v)}(u_2)} \le e^\diffp$ for any $v \in \sphere^{d-1}$ and $u_1,u_2 \in \r^d$ with $\ltwo{u_1} = \ltwo{u_2}$
    \end{enumerate}
\end{lemma}

\begin{proof}
    Given $\R$, we define $\R'$ as follows. First, sample a random rotation matrix $U \in \r^{d \times d}$ where $U^T U = I$, then set
    \begin{equation*}
        \R'(x) = U^T \R(Ux).
    \end{equation*}
    \ificml
    The randomizer $\R'$ satisfies all of our desired properties. We defer the full proof to~\Cref{sec:apdx-pu}.
    \else
    We now prove that $\R'$ satisfies the desired properties. First, note that the privacy of $\R$ immediately implies the same privacy bound for $\R'$. 
    Moreover, we have that 
    $\E[\R'(v)] = \E[U^T \E[\R(Uv)]] = \E[U^T Uv] = v$ as $\R$ is unbiased and $U^T U =I$. For utility, note that
    \begin{align*}
    \err(\R')
        & = \sup_{v \in \sphere^{d-1}} \E_{\R'} \left[ \ltwo{\R'(v) - v}^2 \right] \\  
        & = \sup_{v \in \sphere^{d-1}} \E_{U, \R} \left[ \ltwo{U^T \R(Uv) - U^T Uv}^2 \right] \\
        & = \sup_{v \in \sphere^{d-1}} \E_{U, \R} \left[ \ltwo{\R(Uv) - Uv}^2 \right] \\
        & = \sup_{v \in \sphere^{d-1}, U} \E_{\R} \left[ \ltwo{\R(Uv) - Uv}^2 \right] \\
        & \le \sup_{v \in \sphere^{d-1}} \E_{\R} \left[ \ltwo{\R(v) - v}^2 \right] \\
        & = \err(\R).
    \end{align*}
    
    Finally, we show that the distributions of $\R'(v_1)$ and $\R'(v_2)$ are the same up to rotations. Indeed, let $V \in \r^{d \times d}$ be a rotation matrix such that $v_1 = Vv_2$. We have that $\R'(v_2)= U^T \R(Uv_2)$, which can also be written as
    $\R'(v_2) \eqd (UV)^T \R(UVv_2) \eqd V^T U^T \R(Uv_1) \eqd V^T \R'(v_1)$ as $UV$ is also a random rotation matrix.
    
    Now we prove the final property. Assume towards a contradiction there is $v_1 \in \sphere^{d-1}$ and $u_1,u_2 \in \r^d$ with $\ltwo{u_1} = \ltwo{u_2}$ such that $\frac{\pdf_{\R'(v_1)}(u_1)}{\pdf_{\R'(v_1)}(u_2)} > e^\diffp$. We will show that this implies that $\R'$ is not $\diffp$-DP. In the proof above we showed that $\R'(v_2) \eqd V^T \R'(v_1)$ for $v_1 = Vv_2$. Therefore for $V$ such that $V^T u_1 = u_2$ we get that $\pdf_{\R'(v_2)}(u_2) = \pdf_{\R'(v_1)}(u_1)$ which implies that 
    \begin{equation*}
        \frac{\pdf_{\R'(v_2)}(u_2)}{\pdf_{\R'(v_1)}(u_2)} > e^\diffp,
    \end{equation*}
    which is a contradiction.
    \fi
\end{proof}

\Cref{lemma:symmetry} implies that we can restrict our attention to algorithms have the same density for all inputs up to rotations and hence allows to study their behavior for a single input. Moreover, as we show in the following lemma, given a randomizer that works for a single input, we can extend it to achieve the same error for all inputs.
To facilitate notation, we say that a density $\pdf: \r^d \to \r_+$ is $\diffp$-indistinguishable 
if $\frac{\pdf'(u_1)}{\pdf'(u_2)} \le e^\diffp$ for all $u_1,u_2 \in \r^d$ such that $\ltwo{u_1} = \ltwo{u_2}$. 
\begin{lemma}
\label{lemma:extension}
    Fix $v_0 = e_1 \in \sphere^{d-1}$.
    Let $\pdf: \r^d \to \r_+$ be an $\diffp$-indistinguishable density function with corresponding random variable $\R $ such that that $\E[\R] = e_1$.
    There exists an $\diffp$-DP local randomizer $\R'$ such that $\err(\R',v) = \E[\ltwo{\R - e_1}^2]$ and $\E[\R'(v)] = v$ for all $v \in \sphere^{d-1}$.
\end{lemma}
\begin{proof}
    The proof is similar to the proof of~\Cref{lemma:symmetry}. For any $v \in \sphere^{d-1}$, we let $U(v) \in \r^{d \times d}$ be an orthonormal matrix such that $v_0 = U(v) v$. Then, following~\Cref{lemma:symmetry}, we define $\R'(v) = U^T(v) \R$. The claim immediately follows.
\end{proof}

\Cref{lemma:symmetry} and \Cref{lemma:extension} imply that we only need to study the behavior of randomizer for a fixed input.
Henceforth, we will fix the input to $v=e_1$ 
and investigate properties of the density given $e_1$.

Given $v=(v_1,v_2,\dots,v_d)$ we define its reflection to be $v^- = (v_1,-v_2,\dots,-v_d)$. The next lemma shows that we can assume that the densities at $v$ and $v^-$ are equal for some optimal algorithm.

\begin{lemma}[Reflection Symmetry]
\label{lemma:pairs-equal}
    Let $\pdf: \r^d \to \r_+$ be an $\diffp$-indistinguishable density function with corresponding random variable $\R $ such that that $\E[\R] = e_1$.
    There is $\pdf': \r^d \to \r_+$  with corresponding random variable $\R'$ that satisfies the same properties such that $\err(\R') \le \err(\R)$ and $\pdf'(u) = \pdf'(u^-)$ for all $u \in \r^d$.
\end{lemma}

\begin{proof}
    We define $\pdf'(u) = \frac{\pdf(u) + \pdf(u^-)}{2}$ for all $u \in \sphere^{d-1}$. First, it is immediate to see that $\pdf(u) = \pdf(u^-)$ for all $u \in \r^d$. Moreover, we have
    \begin{align*}
    \frac{\pdf'(u_1)}{\pdf'(u_2)}
          & = \frac{\pdf(u_1) + \pdf(u_1^-)}{\pdf(u_2) + \pdf(u_2^-)} \\
           & \le \max \left(\frac{\pdf(u_1)}{\pdf(u_2)}, \frac{\pdf(u_1^-)}{\pdf(u_2^-)} \right)
           \le e^\diffp.
    \end{align*}
    Note also that $\E[\R'] = e_1$ since the marginal distribution of the first coordinate in the output did not change and it is clear that for other coordinates the expectation is zero as $u + u^- = c \cdot e_1$ for any $u \in \r^d$. Finally, note that $\err(\R',e_1) = \err(\R,e_1)$ since $\ltwo{u-e_1} = \ltwo{u^- - e_1}$ for all $u \in \r^d$.
\end{proof}

We also have the following lemma which shows that the optimal density $\pdf$ outputs vectors on a sphere with some fixed radius.
\begin{lemma}
\label{lemma:fixed-radius}
    Let $\pdf: \r^d \to \r_+$ be an $\diffp$-indistinguishable density function with corresponding random variable $\R $ such that that $\E[\R] = e_1$.
    For any $\tau >0$,
    there exists an $\diffp$-indistinguishable density $\pdf': \r^d \to \r_+$  with corresponding random variable $\R'$ such that $\ltwo{\R'} = C$ for some $C>0$, $\E[\R']=e_1$ and $\err(\R') \le \err(\R) + \tau$.
\end{lemma}
\begin{proof}
    By \Cref{lemma:pairs-equal}, we can assume without loss of generality that $\R$ satisfies reflection symmetry, that is $\pdf(u) = \pdf(u^-)$. 
    We think of the density $\pdf$ as first sampling a radius $R$ then sampling a vector $u \in \r^d$ of radius $R$. We also assume that $\R$ has bounded radius; otherwise as $\err(\R) = \E[\ltwo{\R^2}] - 1$ is bounded, we can project the output of $\R$ to some large radius $R_\tau>0$ while increasing the error by at most $\tau >0$ for any $\tau$. Similarly, we can assume that the output has radius at least $r_{\tau}$ while increasing the error by at most $\tau$.
    Let $\pdf_{R}$ denote the distribution of the radius, and $f_{u\mid R=r}$ be the conditional distribution of the output given the radius is $r$. In this terminology $\E[\R]=e_1$ implies that 
    \begin{align*}
    \err(\R,e_1)
        &= \E[\|\R - e_1\|_2^2] \\
        &=  \E[\|\R\|_2^2 + \|e_1\|_2^2 - 2\langle \R, e_1\rangle] \\&= \E[\ltwo{\R}^2] - 1.
    \end{align*}
    For the purpose of finding the optimal algorithm, we need $\R$ that minimizes $\E[\ltwo{\R}^2]$. Denote $W_r = \E[\<\R,e_1\> \mid R=r] $ and set
    \begin{align*}
    C_{\max} = \sup_{r\in [r_{\tau}, R_{\tau}]} \frac{W_r}{r} .
    \end{align*}
    Noting that $\E[\<\R,e_1\>]=1$, we have
    \begin{align*}
    \E[\ltwo{\R}^2]
        & = \frac{\E[\ltwo{\R}^2]}{\E[\<\R,e_1\>]^2} \\
        & = \frac{\E[\ltwo{\R}^2]}{\left(\int_{r=0}^\infty f_r W_r dr \right)^2} \\
        & = \frac{\E[\ltwo{\R}^2]}{\left( \int_{r=0}^\infty f_r r (W_r/r) dr\right)^2} \\ 
        & \ge \frac{\E[\ltwo{\R}^2]}{\left( \int_{r=0}^\infty f_r r C_{\max} dr \right)^2} \\
        & \ge \frac{1}{C_{\max}^2} \frac{\E[\ltwo{\R}^2]}{\E[\ltwo{\R}]^2} 
         \ge  \frac{1}{C_{\max}^2}.
    \end{align*}
    \newcommand{\cmax}{C_{\max}}
    \newcommand{\rmax}{r_{\max}}
    \newcommand{\fmax}{f_{\max}}
    \newcommand{\Rmax}{\R_{\max}}

    Now consider $\rmax >0$ that has $C_{\max} = W_{\rmax}/\rmax$; $\rmax$ exists as $\R$ has outputs in $[r_{\tau}, R_{\tau}]$. Let $\fmax$ denote the conditional distribution of $\R$ given that $R=\rmax$ and let $\Rmax$ denote the corresponding randomizer. We define a new randomizer $\R'$ as follows
    \begin{equation*}
        \R' = \frac{1}{\rmax \cmax} \Rmax,
    \end{equation*}
    with corresponding density $\pdf'$.
    Note that $\pdf'$ is $\diffp$-indistiguishable as $\pdf$ is $\diffp$-indistiguishable and the conditional distributions given different radii are disjoint which implies $\fmax$ is $\diffp$-indistiguishable. Moreover $\fmax(u) = \fmax(u^-)$ which implies that $\E[\R'] = \frac{1}{\rmax \cmax} \E[R \mid R=\rmax] = e_1$. Finally, note that $\R'$ satisfies
    \begin{align*}
    \E[\ltwo{\R'}^2]
        & = \frac{1}{\rmax^2 \cmax^2} \E[\ltwo{\Rmax}^2] \\
        & = \frac{1}{\rmax^2 \cmax^2} \E[\ltwo{\R}^2 \mid R = \rmax] \\
        & = \frac{1}{\cmax^2}
         \le \E[\ltwo{\R}^2].
    \end{align*}
    The claim follows.
\end{proof}

Before we present our main proposition which formulates the linear program that finds the optimal minimizer, we need the following key property which allows to describe the privacy guarantee as a linear constraint. We remark that such a lemma can easily be proven for deletion DP, so that our results would extend to that definition. 
\begin{lemma}
\label{lemma:ref-repl-DP}
     Let $\R : \sphere^{d-1} \to \r^d $ be an $\diffp$-DP local randomizer. 
     There is $\rho: \r^d \to \r_+ $ such that for all $v \in \sphere^{d-1}$ and $u \in \r$
     \begin{equation*}
         e^{-\diffp/2} \le \frac{\pdf_{\R(v)}(u)}{\rho(u) } \le e^{\diffp/2}.
     \end{equation*}
     Moreover, if $\R$ satisfies the properties of~\Cref{lemma:symmetry} (invariance) then $\rho(u_1) = \rho(u_2)$ for $\ltwo{u_1} = \ltwo{u_2}$.
\end{lemma}

\begin{proof}
\ificml
    Define 
    \[
    \rho(u) = \sqrt{\inf_{v \in \sphere^{d-1}} \pdf_{\R(v)}(u) \cdot \sup_{v \in \sphere^{d-1}} \pdf_{\R(v)}(u) }.\] Note that for all $v \in \sphere^{d-1}$,
\begin{flalign*}
    &\frac{\pdf_{\R(v)}(u)}{\sqrt{\inf_{v \in \sphere^{d-1}} \pdf_{\R(v)}(u) \cdot \sup_{v \in \sphere^{d-1}} \pdf_{\R(v)}(u) }}& \\ &\;\;\;= \sqrt{\frac{\pdf_{\R(v)}(u)}{{\inf_{v \in \sphere^{d-1}} \pdf_{\R(v)}(u)}}}
        \sqrt{\frac{\pdf_{\R(v)}(u)}{{\sup_{v \in \sphere^{d-1}} \pdf_{\R(v)}(u)}}} \le e^{\diffp/2}. 
    \end{flalign*}
\else 
    Define $\rho(u) = \sqrt{\inf_{v \in \sphere^{d-1}} \pdf_{\R(v)}(u) \cdot \sup_{v \in \sphere^{d-1}} \pdf_{\R(v)}(u) }$. Note that for all $v \in \sphere^{d-1}$,
\begin{align*}
    \frac{\pdf_{\R(v)}(u)}{\sqrt{\inf_{v \in \sphere^{d-1}} \pdf_{\R(v)}(u) \cdot \sup_{v \in \sphere^{d-1}} \pdf_{\R(v)}(u) }}     & = \sqrt{\frac{\pdf_{\R(v)}(u)}{{\inf_{v \in \sphere^{d-1}} \pdf_{\R(v)}(u)}}}
        \sqrt{\frac{\pdf_{\R(v)}(u)}{{\sup_{v \in \sphere^{d-1}} \pdf_{\R(v)}(u)}}}
        \le e^{\diffp/2}. 
    \end{align*}
    \fi
    The second direction follows similarly. The second part of the claim follows as for any $u_1,u_2 \in \r^d$ such that $\ltwo{u_1}=\ltwo{u_2}$, if $\pdf_{R(v_1)}(u_1) = t$ for
    any mechanism that satisfies the properties of~\Cref{lemma:symmetry} then there is $v_2$ such that $\pdf_{R(v_2)}(u_2) = t$. The definition of $\rho$ now implies that $\rho(u_1) = \rho(u_2)$.
\end{proof}

We are now ready to present our main step towards proving the optimality result. The following proposition formulates the problem of finding the optimal algorithm as a linear program. As a result, we show that there is an optimal algorithm whose density function has at most two different probabilities.
\begin{proposition}
\label{lemma:two-prob}
    Let $\R : \sphere^{d-1} \to \r^d $ be an $\diffp$-DP local randomizer such that $\E[\R(v)] =v$ for all $v \in \sphere^{d-1}$. For any $\tau>0$, there exist constants $C, p >0$ and an $\diffp$-DP local randomizer $\R' : \sphere^{d-1} \to C \cdot \sphere^{d-1} $ such that $\E[\R'(v)] = v$ for all $v \in \sphere^{d-1}$, $\err(\R') \le \err(\R) + \tau$, $\pdf_{\R'(v)}(u) = \pdf_{\R'(v)}(u^-)$ ,  and $\pdf_{\R'(v)}(u) \in \{e^{-\diffp/2}, e^{\diffp/2}\}  p $ for all $u \in C \cdot \sphere^{d-1}$.
\end{proposition}

\begin{proof}
     The proof will proceed by formulating  a linear program which describes the problem of finding the optimal randomizer and then argue that minimizer of this program must satisfy the desired conditions. To this end, first we use the properties of the optimal randomizer from the previous lemmas to simplify the linear program. \Cref{lemma:fixed-radius} implies that there exists an optimal randomizer $\R: \sphere^{d-1} \to C \cdot \sphere^{d-1}  $ for some $C>0$ that is also invariant under rotations (satisfies the conclusions of~\Cref{lemma:symmetry}).  Moreover, \Cref{lemma:ref-repl-DP} implies that the density function $\pdf_{R(v)}$ has for some $p>0$
    \begin{equation*}
        e^{-\diffp/2} p \le \pdf_{R(v)}(u) \le e^{\diffp/2} p.
    \end{equation*}
    
    Adding the requirement of unbiasedness, and noticing that for such algorithms the error is $C^2 - 1$, this results in the following minimization problem where the variables are $C$ and the density functions $\pdf_v : C \sphere^{d-1} \to \r_+$ for all $v \in \sphere^{d-1}$
    \begin{align*}
        & \arg \min_{C, \pdf_v:C \sphere^{d-1} \to \r_+  \text{for all } v \in \sphere^d }
               C \quad \quad  \label{eq:lp1} \tag{A}\\
        & \subjectto \\
        & \quad \quad  \quad  \quad  \quad e^{-\diffp/2} p \le \pdf_v(u) \le e^{\diffp/2} p, \quad v \in \sphere^{d-1}, ~ u \in C \sphere^{d-1} \\
        & \quad \quad  \quad  \quad  \quad \int_{C \sphere^{d-1}} \pdf_v(u) u du = v, \quad v \in \sphere^{d-1} \\
        & \quad \quad  \quad  \quad  \quad \int_{C \sphere^{d-1}} \pdf_v(u) du =1, \quad v \in \sphere^{d-1}
    \end{align*}
    \Cref{lemma:symmetry} and~\Cref{lemma:extension} also show that the optimal algorithm is invariant under rotations, and that we only need to find the output distribution $\pdf$ with respect to a fixed input $v = e_1$. Moreover,  \Cref{lemma:pairs-equal} says that can assume that $\pdf_{e_1}(u) = \pdf_{e_1}(u^-)$ for all $u$. We also work now with the normalized algorithm $\hat \R(v) = \R(v)/C$ (that is, the output on the unit sphere). Note that for $\hat \R$ we have $\E[\hat \R(v)] = v/C$. Denoting $\alpha = 1/C$, this results in the following linear program (LP)
    \begin{align*}
        & \arg \max_{\alpha, p, \pdf_{e_1}: \sphere^{d-1} \to \r_+  }
               \alpha \quad \quad  \label{eq:lp2} \tag{B} \\
        & \subjectto \\
        & \quad \quad  \quad  \quad  \quad e^{-\diffp/2} p \le \pdf_{e_1}(u) \le e^{\diffp/2} p, \quad  u \in \sphere^{d-1} \\
        & \quad \quad  \quad  \quad  \quad \pdf_{e_1}(u) =  \pdf_{e_1}(u^-), \quad  u \in \sphere^{d-1} \\
        & \quad \quad  \quad  \quad  \quad \int_{C \sphere^{d-1}} \pdf_{e_1}(u) u du = \alpha e_1, \\
        & \quad \quad  \quad  \quad  \quad \int_{C \sphere^{d-1}} \pdf_{e_1}(u) du =1.
    \end{align*} 
    We need to show that most of the inequality constraints $e^{-\diffp/2} p \le \pdf_{e_1}(u) \le e^{\diffp/2} p$ must be tight at one of the two extremes. To this end, we approximate the LP~\eqref{eq:lp2} using a finite number of variables by discretizing the density function $f_{e_1}$. We assume we have a $\delta/2$-cover $S = \{u_1, \dots, u_K \}$ of $\sphere^{d-1}$. We assume without loss of generality that if $u_i \in S$ then $u_i^- \in S$ and we also write $S = S_0 \cup S_1$ where $S_0 = S_1^-$ and $S_0 \cap S_1 = \emptyset$. 
    Let $B_i = \{w \in \sphere^{d-1}: \ltwo{w - u_i} \leq \ltwo{w-u_j} \} $, $V_i = \int_{u \in B_i} 1 du$, and $\bar u_i = \E_{U \sim \uniform(\sphere^{d-1})}[U \mid U \in B_i]$. Now we limit our linear program to density functions that are constant over each $B_i$, resulting in the following LP
    \begin{align*}
        & \arg \max_{\alpha, \pdf_{e_1}: \sphere^{d-1} \to \r_+  }
               \alpha \quad \quad \label{eq:lp3} \tag{C} \\
        & \subjectto \\
        & \quad \quad  \quad  \quad  \quad e^{-\diffp/2} p\le \pdf_{e_1}(u) \le e^{\diffp/2} p, \quad  u \in S_0 \\
        & \quad \quad  \quad  \quad  \quad \sum_{u \in S_0} \pdf_{e_1}(u)V_i \bar u_i + \sum_{u \in S_1} \pdf_{e_1}(u^-)V_i \bar u_i = \alpha e_1, \\
        & \quad \quad  \quad  \quad  \quad  \sum_{u \in S_0} \pdf_{e_1}(u) V_i + \sum_{u \in S_1} \pdf_{e_1}(u^-) V_i  =1. 
    \end{align*} 
    
     Let $\alpha^\star_1$ and $\alpha^\star_2$ denote the maximal values of~\eqref{eq:lp2} and~\eqref{eq:lp3}, respectively. Each solution to~\eqref{eq:lp3} is also a solution to~\eqref{eq:lp2} hence $\alpha^\star_1 \ge \alpha^\star_2$. Moreover, given $\delta>0$, let $\pdf$ be a  solution of~\eqref{eq:lp2} that obtains $\alpha \ge \alpha^\star_1 - \delta$ and let $\R$ be the corresponding randomizer. 
     We can now define a solution for the discrete program~\eqref{eq:lp3} by setting for $u \in B_i$,
     \begin{equation*}
         \hat \pdf(u) = \frac{1}{V_i} \int_{w \in B_i} \pdf_{e_1}(w) dw
     \end{equation*}
      Equivalently, we can define $\hat \R$ as follows: first run $\R$ to get $u$ and find $B_i$ such that $u \in B_i$. Then return a vector uniformly at random from $B_i$.
     Note that $\hat \pdf$ clearly satisfies the first and third constraints in~\eqref{eq:lp3}. As for the second constraint, it follows since $\hat \pdf_{e_1}(u) = \hat \pdf_{e_1}(u^-)$ which implies that  $\sum_{u \in S} \pdf_{e_1}(u)V_i \bar u_i = \hat \alpha v$ for some $\hat \alpha >0$. It remains to show that $\hat \alpha \ge \alpha^\star_1 - 2 \delta$. The above representation of $\hat \R$ shows that $\|\E[\hat \R - \R]\| \le \delta$ and therefore we have $\hat \alpha \ge \alpha^\star_1 - 2\delta$.

      To finish the proof, it remains to show that the discrete LP~\eqref{eq:lp3} has a solution that satisfies the desired properties. 
      Note that as this is a linear program with $K$ variables and $2K+d+2$ constraints, $K$ linearly independent constraints must be satisfied~\cite[theorem 2.3]{BertsimasTs97}, 
      which shows that for at least $K-d-2$ of the sets $B_i$ we have $\pdf(u_i) \in \{e^{\diffp},e^{-\diffp}\}  p$.
      
      To finish the proof, we need to manipulate the probabilities for the remaining $d-2$ sets to satisfy our desired requirements. As these sets have small probability, this does not change the accuracy by much and we just need to do this manipulation carefully so as to preserve reflection symmetry and unbiasedness. The full details are tedious and we present them in~\Cref{sec:apdx-two-prop-proof}.
\end{proof}

Given the previous lemmas, we are now ready to finish the proof of~\Cref{thm:opt-repl}.
\begin{proof}
    
    \newcommand{\rmax}{r_{\max}}
    \newcommand{\fmax}{f_{\max}}
    
    Fix $\tau >0$ and an unbiased $\diffp$-DP local randomizer $\R\opt$. \Cref{lemma:two-prob} shows that there exists $\R: \sphere^{d-1} \to \rmax \sphere^{d-1}$ that is $\diffp$-DP, unbiased, reflection symmetric ($\pdf_{\R(e_1)}(u) = \pdf_{\R(e_1)}(u^-)$ for all $u$), and satisfies  $\pdf_{\R(e_1)}(u) \in \{e^{-\diffp/2},e^{\diffp/2}\} p$. Morevoer  $\err(\R) \le \err(\R\opt) + \tau$. We will transform $\R$ into an instance of $\pu$ while maintaining the same error as $\R$.

    To this end, if $\R$ is an instance of $\pu$ then we are done. Otherwise 
    let $\pdf = \pdf_{\R(e_1)}$, $S_0(t) = \{u: \pdf(u) = p e^{\diffp/2}, \<u,e_1\> \le t\}$ and $S_1(t) = \{u: \pdf(u) = p e^{-\diffp/2}, \<u,e_1\> \ge t\}$.
    Consider $t \in [-1,1]$ that solves the following minimization problem
    \begin{equation*}
    \begin{split}
    \minimize_{t \in [-1,1]} ~ & \int_{S_0(t)} \pdf(u)du + \int_{S_1(t)} \pdf(u)du 
     \end{split}
    \end{equation*}
    Let $p\opt$ be the value of the above minimization problem and $t\opt$ the corresponding minimizer. Let $p_0 = \int_{S_0(t\opt)} \pdf(u)du $ and $p_1 = \int_{S_1(t\opt)} \pdf(u)du $. Assume without loss of generality that $p_0 \le p_1$ (the other direction follows from identical arguments). Let $\hat S \subseteq S_1(t\opt)$ be such that $\hat S = \hat S^-$ and  $\int_{\hat S} \pdf(u) du = p_0$. We define $\tilde f$ by swapping the probabilities on $\hat S$ and $S_0(t\opt)$, that is,
    \begin{equation*}
    \tilde f(u) = 
        \begin{cases}
         \pdf(u) & \text{if } u \notin S_0(t\opt) \cup \hat S \\
         p e^{-\diffp/2} & \text{if } u \in S_0(t\opt) \\
         p e^{\diffp/2} & \text{if } u \in \hat S         \end{cases}
    \end{equation*}    
    Clearly $\tilde \pdf$ still satisfies all of our desired properties and has $\E_{U \sim \tilde \pdf}[\<U,e_1\>] \ge \E_{U \sim \pdf}[\<U,e_1\>]$ as we have that $\<u_1,e_1\> \ge \<u_2,e_1\>$ for $u_1 \in \hat S$ and $u_2 \in S_0(t\opt)$. Note also that $\tilde \pdf(u) = p e^{-\diffp/2}$ for $u$ such that $\<u,e_1\> \le t$. Moreover, for $u$ such that $\<u,e_1\> \ge t$, we have that $\tilde \pdf(u) = p  e^{-\diffp/2}$ only if $u \in B \defeq S_1 \setminus \hat S$. Let $\delta$ be such that the set $A = \{u: t\opt \le \<u,e_1\> \le t\opt + \delta \}$ has $\int_{u \in A} \tilde \pdf(u) du = \int_{u \in B} \tilde \pdf(u) du$. We now define
    \begin{equation*}
    \hat f(u) = 
        \begin{cases}
         p e^{\diffp/2} & \text{if } \< u,e_1\> \ge t\opt + \delta \\
         p e^{-\diffp/2} & \text{if } \< u,e_1\> \le t\opt + \delta     
         \end{cases}
    \end{equation*}
    Clearly, $\hat f(u) $ is an instance of $\pu$. Now we prove that it satisfies all of our desired properties. First, note that we can write $\hat f$ as 
    \begin{equation*}
    \hat f(u) = 
        \begin{cases}
         \tilde \pdf(u) & \text{if } u \notin A \cup  B \\
         \tilde \pdf(u) & \text{if } u \in A \cap B \\
         p e^{-\diffp/2} & \text{if } u \in A \setminus B \\
         p e^{\diffp/2} & \text{if } u \in B \setminus A    
         \end{cases}
    \end{equation*}     
    This implies that $\int_{\sphere^{d-1}} \hat \pdf du = 1$. Moreover, $\hat f$ is $\diffp$-indistiguishable by definition. Finally, note that $\E_{U \sim \hat \pdf}[\<U,e_1\>] \ge \E_{U \sim \tilde \pdf}[\<U,e_1\>]$ as $\<u_1,e_1\> \ge \<u_2,e_1\>$ for $u_1 \in B \setminus A$ and $u_2 \in A \setminus B$. Let $\hat \R$ be the randomizer that corresponds to $\hat f$. We define $\R' = \frac{1}{\E_{U \sim \hat \pdf}[\<U,e_1\>]} \hat \R$. We have that $\E[\R'] = e_1$ and that 
    \ificml
    \begin{align*}
    \err(\R') 
        & = \frac{1}{\E_{U \sim \hat \pdf}[\<U,e_1\>]^2} - 1 \\
        & \le \frac{1}{\E_{U \sim \pdf}[\<U,e_1\>]^2} - 1   
        = \err(\R).
    \end{align*}
    \else
    \begin{align*}
    \err(\R') 
        & = \frac{1}{\E_{U \sim \hat \pdf}[\<U,e_1\>]^2} - 1 \\
        & \le \frac{1}{\E_{U \sim \pdf}[\<U,e_1\>]^2} - 1 \\
        &  = \err(\R).
    \end{align*}
    \fi
    As $\R'$ is an instance of $\pu$, the claim follows.
%
    %
%
\end{proof}

\section{$\pug$: an optimal algorithm based on Gaussian distribution}
\label{sec:pug}
In this section, we develop a new variant of $\pu$, namely $\pug$, based on the Gaussian distribution. 
$\pug$ essentially provides an easy-to-analyze approximation of the optimal algorithm $\pu$. This enables to efficiently find accurate approximations of the optimal parameters $p^\star$ and $q^\star$. In fact, we show that these parameters are independent of the dimension which is computationally valuable. Moreover, building on $\pug$, we are able to analytically study the constants that characterize the optimal loss.

\ificml
For a Gaussian random variable $U = \normal(0,1/d)$ and input vector $v$, $\pug$ has the following distribution (up to normalization constants) 
\else
The main idea in $\pug$ is to approximate the uniform distribution over the unit sphere using a Gaussian distribution. Roughly, for a Gaussian random variable $U = \normal(0,1/d)$ and input vector $v$, $\pug$ has the following distribution (up to normalization constants) 
\fi
\begin{equation*}
\pug \sim 
        \begin{cases}
            & U \mid \<U,v\> \ge \gamma  \quad \quad  \text{with probability } p \\
            & U \mid \<U,v\> < \gamma \quad \quad  \text{with probability } 1-p
        \end{cases}
\end{equation*}
We present the full details including the normalization constants in~\Cref{alg:puG}. We usually use the notation $\pug(p,q)$ which means applying $\pug$ with $p$ and $\gamma  =  \Phi^{-1}(q)/\sqrt{d}$.

\ificml
\else
\begin{algorithm}
	\caption{$\pug(p,q)$}
	\label{alg:puG}
	\begin{algorithmic}[1]
		\REQUIRE $v \in \sphere^{d-1}$, $q \in [0,1]$, $p \in [0,1]$
		\STATE Draw $z \sim \mathsf{Ber}(p)$
		\STATE Let $U = \normal(0,\sigma^2) $ where $\sigma^2 = 1/d$
		\STATE Set $\gamma = \Phi_{\sigma^2}^{-1}(q) = \sigma \cdot \Phi^{-1}(q)$
		\IF{$z=1$}
		    \STATE Draw $\alpha \sim U \mid U \ge \gamma$
		\ELSE
		     \STATE Draw $\alpha \sim U \mid U < \gamma$
		\ENDIF
		\STATE Draw $V^{\perp} \sim \normal(0, \sigma^2(I - v v^T)$
		\STATE Set $V = \alpha v + V^{\perp}$
		\STATE Calculate
		\begin{equation*}
		    m = \sigma \phi(\gamma/\sigma) \left(\frac{p}{1 - q}  - \frac{1-p}{q} \right)
		\end{equation*}
        \STATE Return $\frac{1}{m} \cdot V$
	\end{algorithmic}
\end{algorithm}
\fi

The following proposition gives the privacy and utility guarantees for $\pug$. 
The r.v. $\alpha$ is defined (see~\Cref{alg:puG}) as $\alpha = \langle U, v\rangle$ where $U$ is drawn from $\pug(v)$. We define $m = \sigma \phi(\gamma/\sigma) \left(\frac{p}{1 - q}  - \frac{1-p}{q} \right)$ with $\sigma^2=1/d$ and $\gamma = \sigma \cdot \Phi^{-1}(q)$. We defer the proof to~\Cref{sec:proof-prop-pug-guarantees}.
\begin{proposition}
\label{prop:pug-guarantees}
    Let $p,q \in [0,1]$ such that $\frac{p}{1-p}\frac{q}{1-q} \le e^\diffp$.
    The algorithm $\pug(p,q)$ is $\diffp$-DP local randomizer. Moreover, it is unbiased and has error
\ificml
\begin{align*}
        \err(\pug(p,q))
        &=  \frac{\E [\alpha^2] + \frac{d-1}{d}}{\E[\alpha]^2} - 1.
    \end{align*}
\else
\begin{equation*}
        \err(\pug(p,q))
        = \E[\ltwo{\pug(v) - v}^2] 
        =  \frac{\E [\alpha^2] + \frac{d-1}{d}}{\E[\alpha]^2} - 1.
    \end{equation*}
    \fi
    Moreover, we have
    \begin{equation*}
        m^2 \cdot \err(\pug(p,q))
        \stackrel{d \to \infty}{\to} 1.
    \end{equation*}
\end{proposition}

\ificml
\begin{figure*}[t]
  \begin{center}
    \begin{tabular}{ccc}
      \begin{overpic}[width=.33\linewidth]{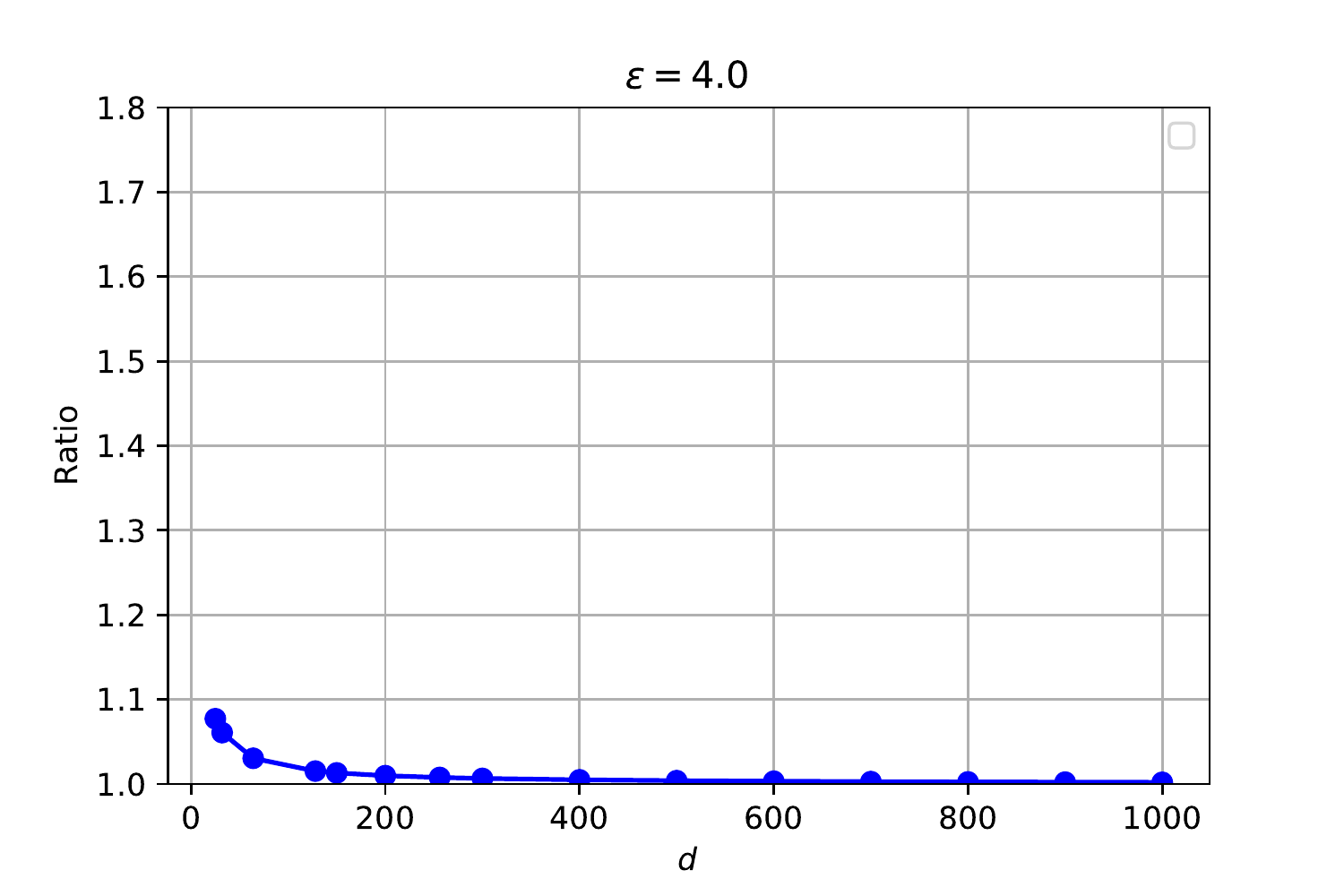}
      \put(48.5,0){
          \tikz{\path[draw=white,fill=white] (0, 0) rectangle (.1,.4);}}
      \put(-2,25){
          \tikz{\path[draw=white,fill=white] (0, 0) rectangle (.35,1);}}
      \put(10,59){
          \tikz{\path[draw=white,fill=white] (0, 0) rectangle (10,.5);}}
      \put(48,0){{\small d}}
        \put(-1,25){
          \rotatebox{90}{{\small Ratio}}}
      \end{overpic} &
      \begin{overpic}[width=.33\linewidth]{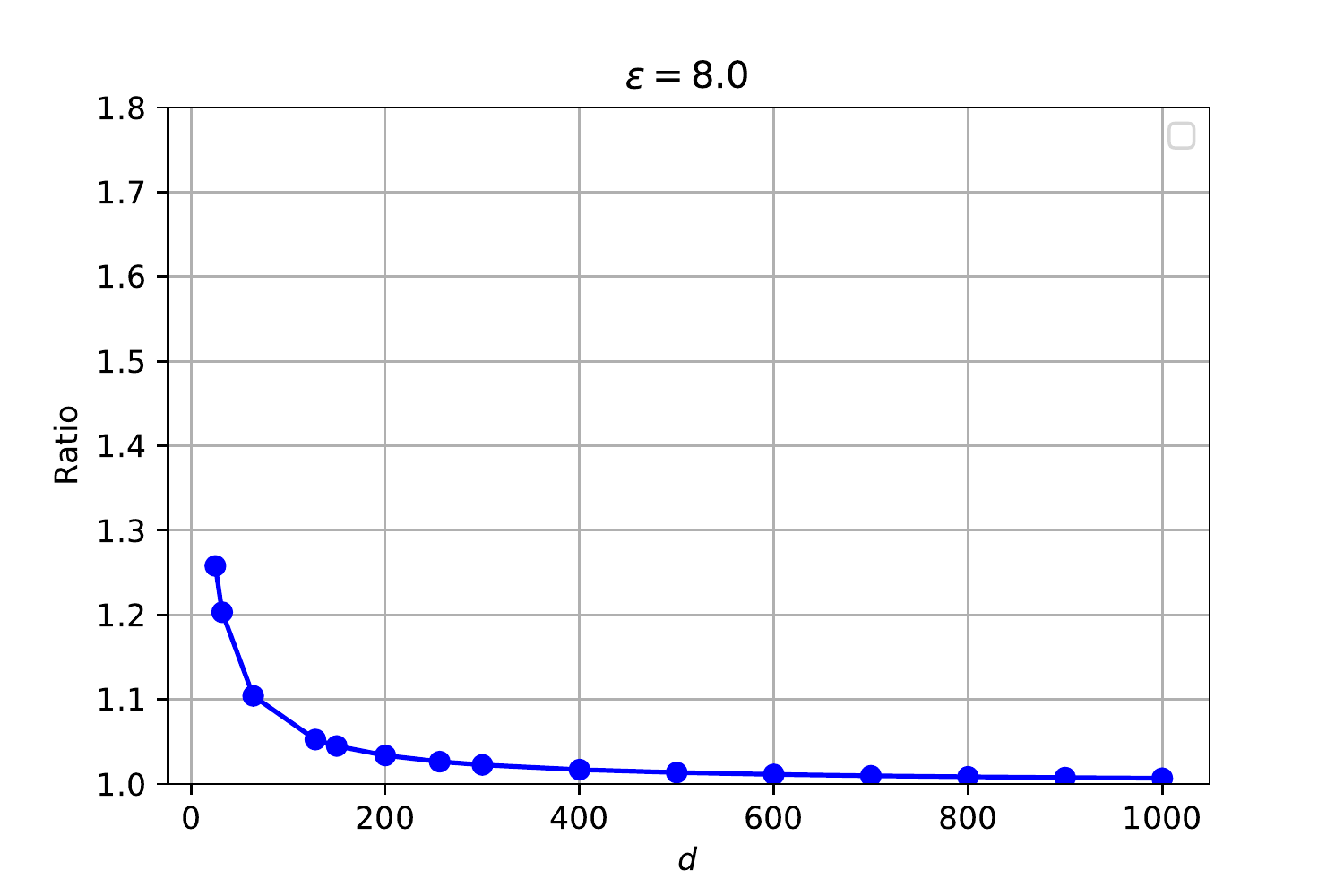}
      \put(48.5,0){
          \tikz{\path[draw=white,fill=white] (0, 0) rectangle (.1,.4);}}
      \put(-2,25){
          \tikz{\path[draw=white,fill=white] (0, 0) rectangle (.35,1);}}
      \put(10,59){
          \tikz{\path[draw=white,fill=white] (0, 0) rectangle (10,.5);}}
      \put(48,0){{\small d}}
        \put(-1,25){
          \rotatebox{90}{{\small Ratio}}}
      \end{overpic}  &
      \begin{overpic}[width=.33\linewidth]{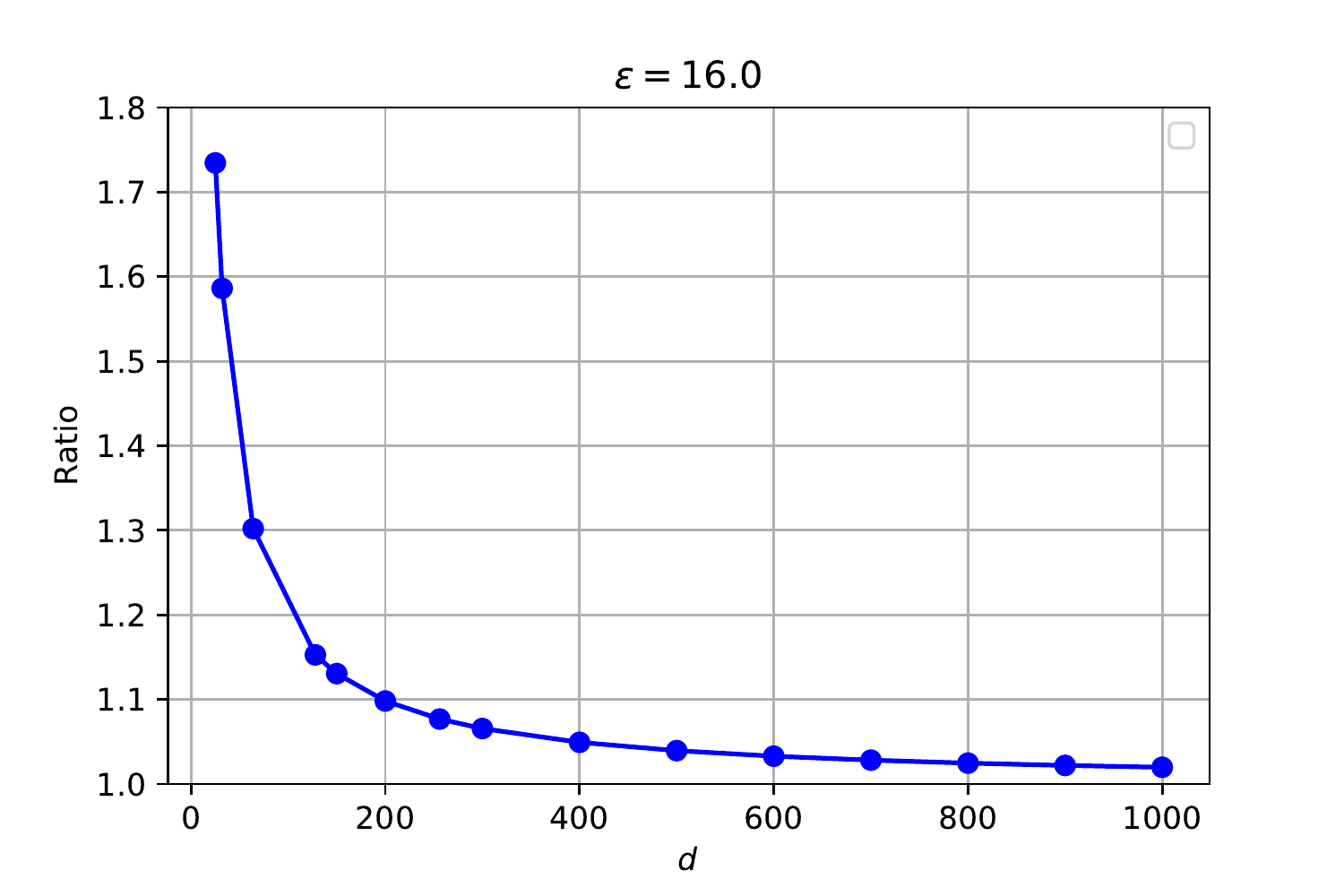}
    \put(48.5,0){
          \tikz{\path[draw=white,fill=white] (0, 0) rectangle (.1,.4);}}
      \put(-2,25){
          \tikz{\path[draw=white,fill=white] (0, 0) rectangle (.35,1);}}
      \put(10,59){
          \tikz{\path[draw=white,fill=white] (0, 0) rectangle (10,.5);}}
      \put(48,0){{\small d}}
        \put(-1,25){
          \rotatebox{90}{{\small Ratio}}}
      \end{overpic}  \\
      (a) & (b) & (c)
    \end{tabular}
    \caption{\label{fig:comp_pu_pug} Ratio of the error of $\pug$ to $\pu$ for (a) $\diffp = 4.0$ (b) $\diffp = 8.0$ (c) $\diffp = 16.0$. We use the same $p$ and $\gamma$ for both algorithms by finding the best $p,q$ that minimize $\pug$.}  \end{center}
\end{figure*}
\fi

\ificml
\begin{figure*}[t]
  \begin{center}
    \begin{tabular}{cc}
      \begin{overpic}[width=.4\linewidth]{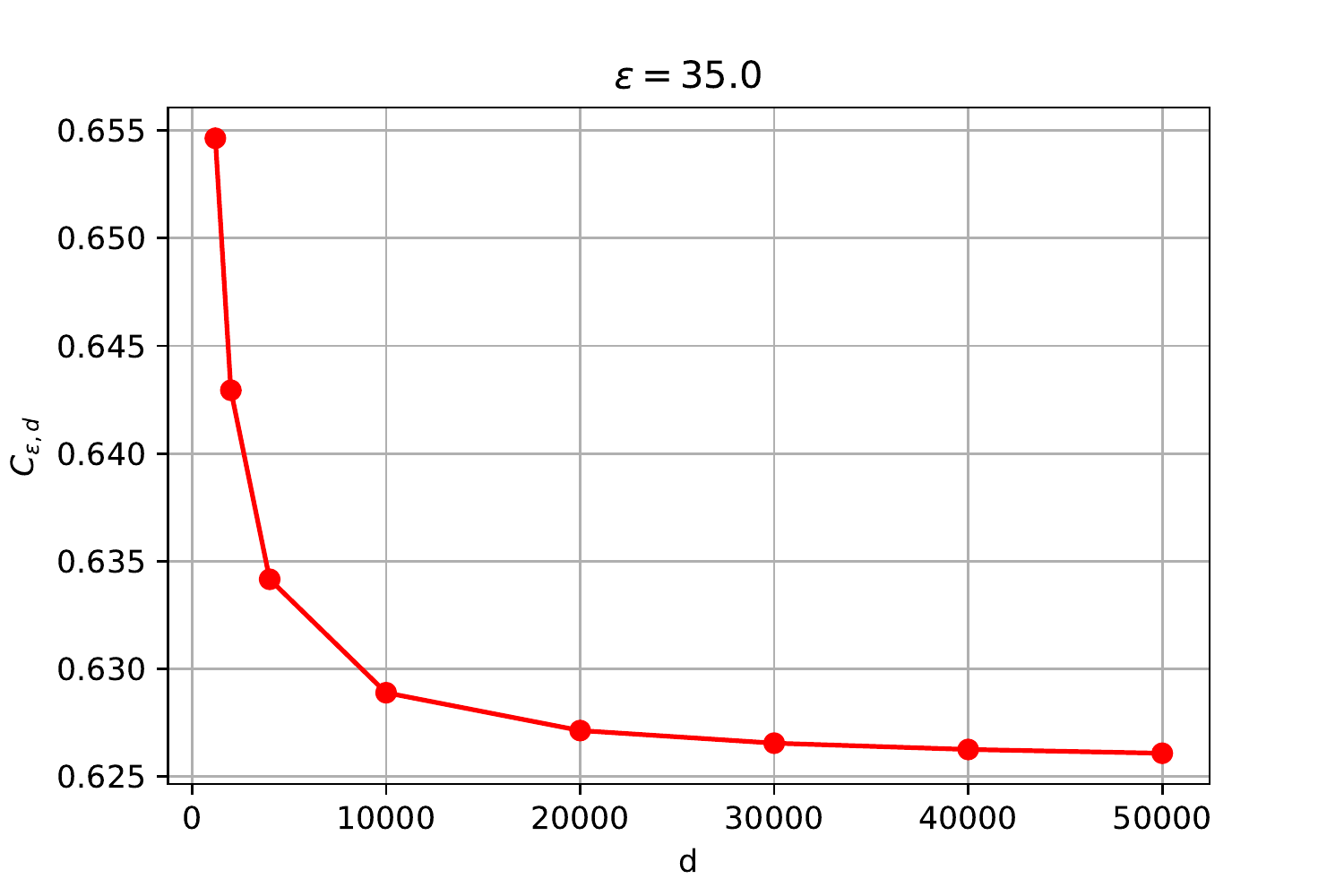}
      \put(48.5,0){
          \tikz{\path[draw=white,fill=white] (0, 0) rectangle (.2,.4);}}
      \put(-2,25){
          \tikz{\path[draw=white,fill=white] (0, 0) rectangle (.3,1);}}
      \put(10,59){
          \tikz{\path[draw=white,fill=white] (0, 0) rectangle (10,.5);}}
      \put(48,0){{d}}
        \put(-3,27){
          \rotatebox{90}{{$C_{\diffp,d}$}}}
      \end{overpic} &
      \begin{overpic}[width=.4\linewidth]{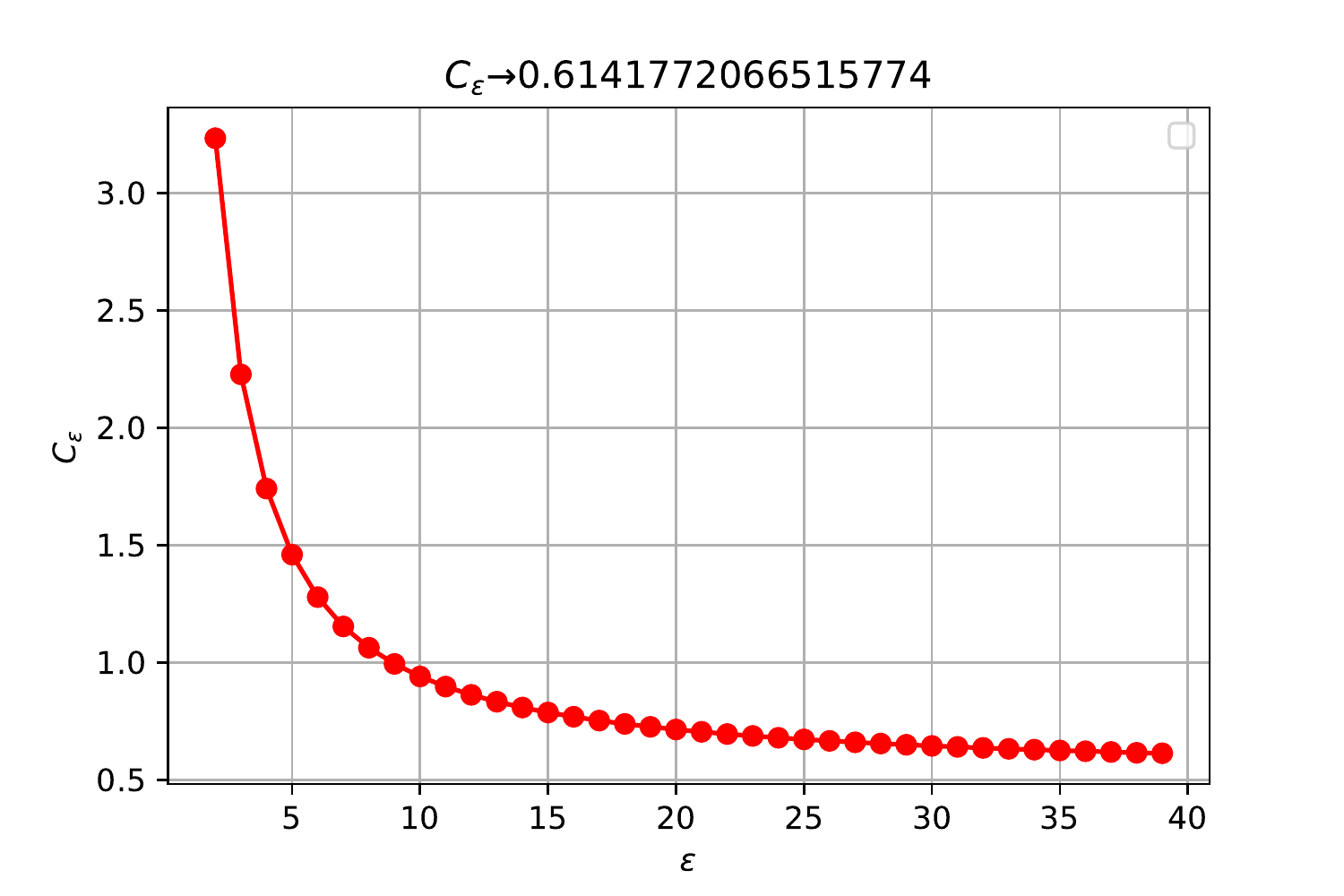}
    \put(48.5,0){
          \tikz{\path[draw=white,fill=white] (0, 0) rectangle (.2,.3);}}
      \put(-2,25){
          \tikz{\path[draw=white,fill=white] (0, 0) rectangle (.5,1);}}
      \put(10,59){
          \tikz{\path[draw=white,fill=white] (0, 0) rectangle (10,.5);}}
      \put(48,0){{$\diffp$}}
        \put(-3,27){
          \rotatebox{90}{{$C_{\diffp}$}}}
      \end{overpic} \\
      (a) & (b) 
    \end{tabular}
    \caption{\label{fig:C-eps} (a) $C_{\diffp,d}$ as a function of $d$ for $\diffp = 35$. (b) $C_\diffp$ as a function of $\diffp$ (we approximate $C_\diffp$ by taking a sufficiently large dimension $d = 5 \cdot 10^4$)}
  \end{center}
  
  \end{figure*}
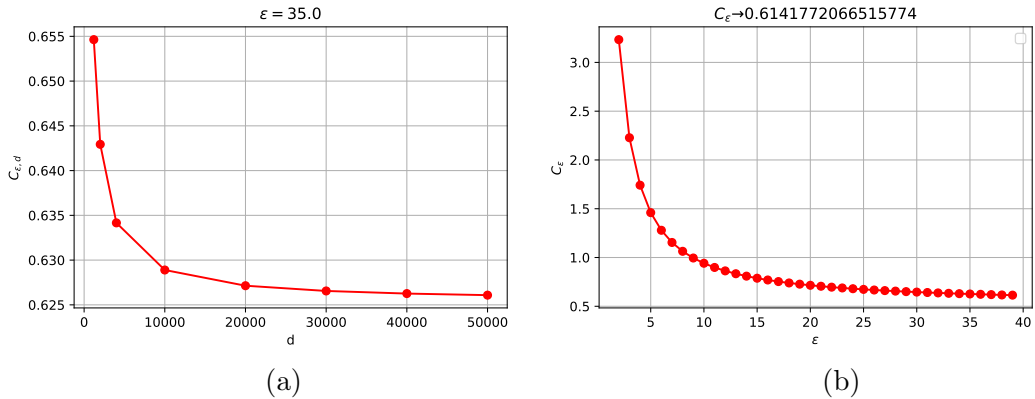
\fi

Now we proceed to analyze the utility guarantees of $\pug$ as compared to $\pu$. To this end, we first define the error obtained by $\pug$ with optimized parameters
\begin{equation*}
    \err^\star_{\diffp,d}(\pug) 
        = \inf_{p,q: \frac{pq}{(1-p)(1-q)} \le e^\diffp} \err(\pug(p,q)).
\end{equation*}
Similarly, we define this quantity for $\pu$
\begin{equation*}
    \err^\star_{\diffp,d}(\pu) 
        = \inf_{p,q: \frac{pq}{(1-p)(1-q)} \le e^\diffp} \err(\pu(p,q)).
\end{equation*}
The following theorem shows that $\pug$ enjoys the same error as $\pu$ up to small factors. 
\ificml We prove the theorem in~\Cref{sec:proof-pug-opt}.
\fi
\begin{theorem}
\label{thm:pug-opt}
    Let $p \in [0,1]$ and $q \in [0,1]$ such that $\pu(p,q)$ is $\diffp$-DP local randomizer. Then $\pug(p,q)$ is also $\diffp$-DP local randomizer and has 
    \begin{align*}
        \err(\pug(p,q)) & \le \err(\pu(p,q))\left( 1  +   O \left( \sqrt{\frac{\diffp + \log d}{d}} \right) \right).
    \end{align*}
    In particular,
    \begin{align*}
        \frac{ \err^\star_{\diffp,d}(\pug)}{ \err^\star_{\diffp,d}(\pu) }  \le 1  +   O \left( \sqrt{\frac{\diffp + \log d}{d}} \right).
    \end{align*}
\end{theorem}

We conduct several experiments that demonstrate that the error of both algorithms is nearly the same as we increase the dimension. We plot the ratio of the error of $\pug$ and $\pu$ (for the same $p$ and $\gamma$) for different epsilons and dimensions in~\Cref{fig:comp_pu_pug}. These plots reaffirm the theoretical results of~\Cref{thm:pug-opt}, that is, the ratio is smaller for large $d$ and small $\diffp$.

\ificml
\else
\begin{figure}[t]
  \begin{center}
    \begin{tabular}{ccc}
      \begin{overpic}[width=.35\columnwidth]{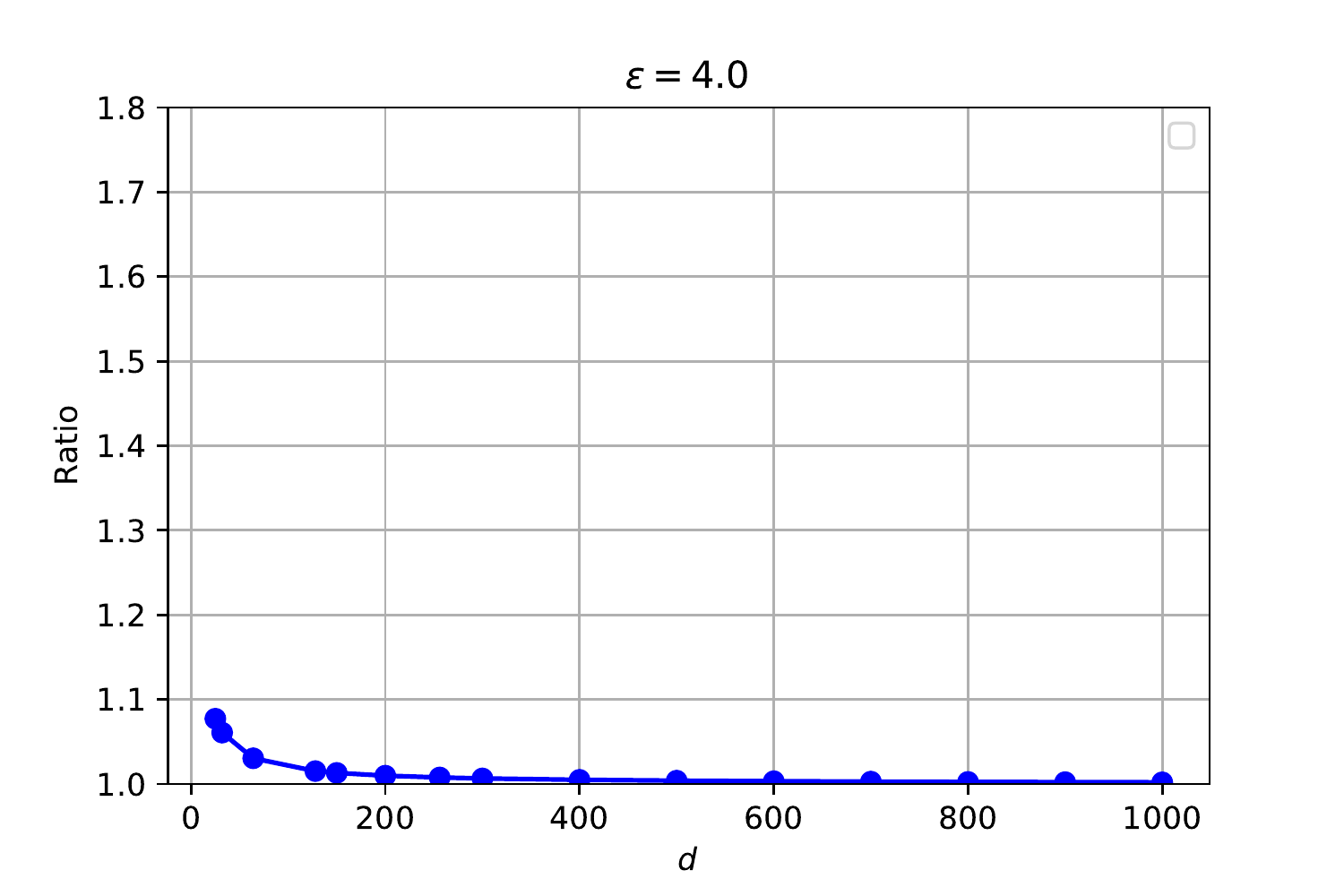}
      \end{overpic} &
      \hspace{-1cm}
      \begin{overpic}[width=.35\columnwidth]{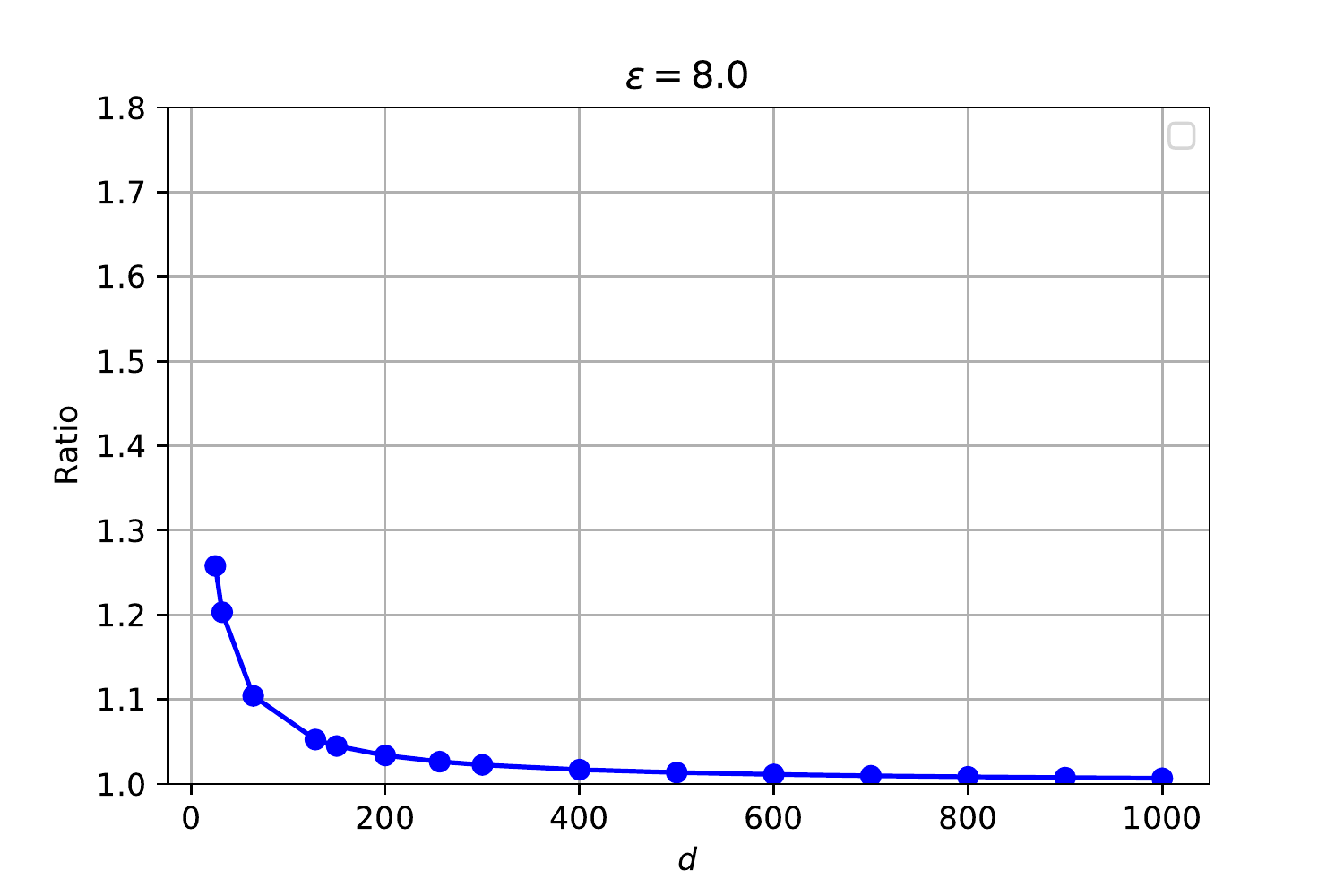}
      \end{overpic}  &
      \hspace{-1cm}
      \begin{overpic}[width=.35\columnwidth]{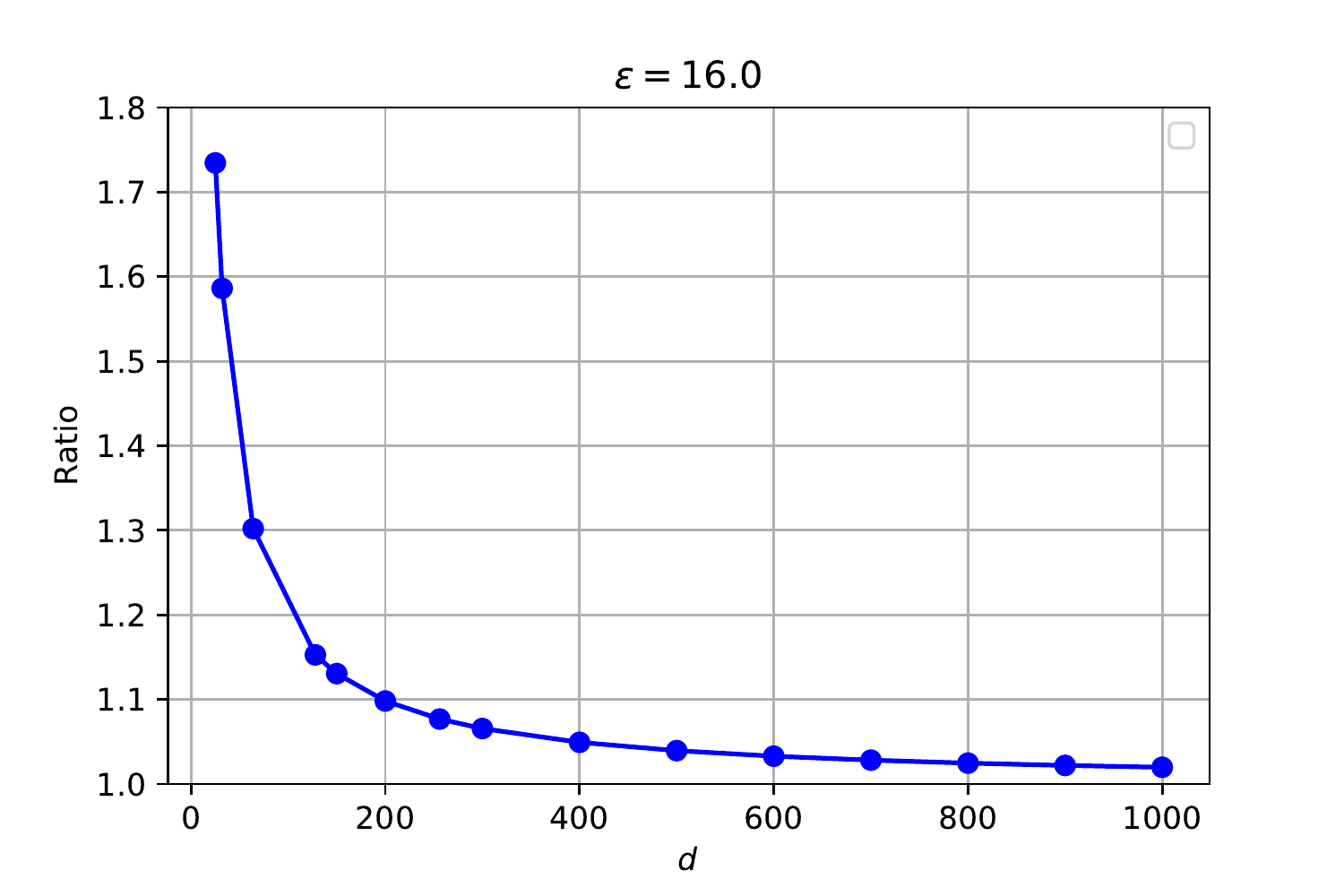}
      \end{overpic}  \\
      (a) & (b) & (c)
    \end{tabular}
    \caption{\label{fig:comp_pu_pug} Ratio of the error of $\pug$ to $\pu$ for (a) $\diffp = 4.0$ (b) $\diffp = 8.0$ (c) $\diffp = 16.0$. We use the same $p$ and $\gamma$ for both algorithms by finding the best $p,q$ that minimize $\pug$.}
  \end{center}

\end{figure}
\fi

\ificml
\else
\begin{proof}
    The privacy proof is straightforward as both algorithms use the same $p$ and $q$ therefore enjoy the same privacy parameter. Moreover, the second part of the claim follows immediately from the first part and the optimality of $\pu$ (\Cref{thm:opt-repl}).
    
    Now we prove the first part of the claim. Let $\gamma_2$ be such that $q = P(W \le \gamma_2)$ where $W \sim \uniform(\sphere_2^{d-1})$. Then we have
    \begin{align*}
    m_2
        & = \E[W \indic{W \ge \gamma_2 }] \frac{p+q-1}{q(1-q)}  \\
    \err_2 & = \err(\pu(p,q))  = \frac{1}{m_2^2}    .
    \end{align*}
    Similarly for $\pug$ we have (see e.g. proof of~\Cref{prop:pug-guarantees})
    \begin{align*}
    m_G
         & = E[\alpha]
           = \E[U \indic{U \ge \gamma_G }] \frac{p+q-1}{q(1-q)} \\
    \err_G & = \err(\pug(p,q)) = \frac{\E[\alpha^2]-1/d}{m_G^2} + \frac{1}{m_G^2}  .
    \end{align*}
    Therefore we have
    \begin{align}
    \label{eq:err-ratio}
    \sqrt{\frac{\err_G}{\err_2}}
        & \le \frac{m_2}{m_G} \sqrt{1 + \E[\alpha^2]} \nonumber \\
        & \le \frac{\E[W \indic{W \ge \gamma_2 }]}{\E[U \indic{U \ge \gamma_G }]} \sqrt{1 + \E[\alpha^2]} \nonumber \\
        & \stackrel{(i)}{\le}  \frac{\E[W \indic{W \ge \gamma_2 }]}{\E[U \indic{U \ge \gamma_G }]}  \left(1 +  O\left( \gamma_G + \sqrt{\frac{\diffp + \log d}{d}}\right) \right), 
    \end{align}
    where inequality $(i)$ follows~\Cref{lemma:bound-sec-moment}. 
    We now upper bound the first term. Note first that for the same $\gamma$ we have
    \begin{align*}
     \frac{\E[W \indic{W \ge \gamma }]}{\E[U \indic{U \ge \gamma}]} 
        & \stackrel{(i)}{\le} 1 + O \left( \frac{ {\sqrt{\log d}}{}}{d^{3/2} \E[U \indic{U \ge \gamma }]} \right) \\
        & \stackrel{(ii)}{\le} 1 + O \left( \frac{ \sqrt{\log d}}{d} \right),
    \end{align*}
    where $(i)$ follows from~\Cref{lemma:bound-diff-exp} and $(ii)$ follows since $\E[U \indic{U \ge \gamma }] \ge \E[U \indic{U \ge 0}] = \E[|U|]/2 \ge \Omega(\sigma) = \Omega(1/\sqrt{d}) $.
    We will need one more step to finish the proof as $\gamma_2$ and $\gamma_G$ are not necessarily equal. 
    We divide to cases. If $\gamma_2 \ge \gamma_G$, then $\E[U \indic{U \ge \gamma_G }] \ge \E[U \indic{U \ge \gamma_2 }]$ and therefore
    \begin{align*}
     \frac{\E[W \indic{W \ge \gamma_2 }]}{\E[U \indic{U \ge \gamma_G }]}
        & \le \frac{\E[W \indic{W \ge \gamma_2 }]}{\E[U \indic{U \ge \gamma_2 }]}.
    \end{align*}
    Similarly, if $\gamma_2 \le \gamma_G$  then $\E[W \indic{W \ge \gamma_2 }] \le \E[W \indic{W \ge \gamma_G }]$ and therefore
    \begin{align*}
     \frac{\E[W \indic{W \ge \gamma_2 }]}{\E[U \indic{U \ge \gamma_G }]}
        & \le \frac{\E[W \indic{W \ge \gamma_G }]}{\E[U \indic{U \ge \gamma_G }]}.
    \end{align*}
    Overall this proves that 
    \begin{align*}
     \frac{\E[W \indic{W \ge \gamma_2 }]}{\E[U \indic{U \ge \gamma_G }]}
        & \le 1 + O \left( \frac{ \sqrt{\log d}}{d} \right).
    \end{align*}
    Putting this back in inequality~\eqref{eq:err-ratio}, we have
    \begin{align*}
    \sqrt{\frac{\err_G}{\err_2}}
        & \le \left( 1 + O \left( \frac{ \sqrt{\log d}}{d} \right) \right)\left(1 +  O\left( \gamma_G + \sqrt{\frac{\diffp + \log d}{d}}\right) \right) \\
        & \le 1 + O \left( \sqrt{\frac{\diffp + \log d}{d}} \right),
    \end{align*}
    where the last inequality follows from~\Cref{lemma:ub-gamma} as $\gamma \le O(\sqrt{{(\diffp+1)}/{d}})$. The claim follows.
\end{proof}
\fi

\subsection{Analytical expression for optimal error}
We wish to understand the constants that characterize the optimal error. To this end, we build on the optimality of $\pug$ and define the quantity $C_{\diffp,d}$ by
\begin{equation*}
    \err^\star_{\diffp,d} (\pug) 
    = C_{\diffp,d} \frac{d}{\diffp}.
\end{equation*}
We show that $C_{\diffp,d} \to C_\diffp$ as $d \to \infty$. Moreover, $C_{\diffp} \to C^\star$ as $\diffp \to \infty$.
We experimentally demonstrate the behavior of $C_{\diffp,d}$ and $C_\diffp$ in~\Cref{fig:C-eps}. These experiments show that $C^\star \approx 0.614$. We remark that as shown in~\cite{FeldmanTa21}, if $C_{\diffp}/C_{k\diffp}$ is close to $1$, then one can get a near optimal algorithm for privacy parameter $k\eps$ by repeating the algorithm for privacy parameter $\eps$ $k$ times. The latter may be more efficient in terms of computation and this motivates understanding how quickly $C_{\eps}$ converges.

\ificml
\else
\begin{figure}[t]
  \begin{center}
    \begin{tabular}{cc}
      \begin{overpic}[width=.45\columnwidth]{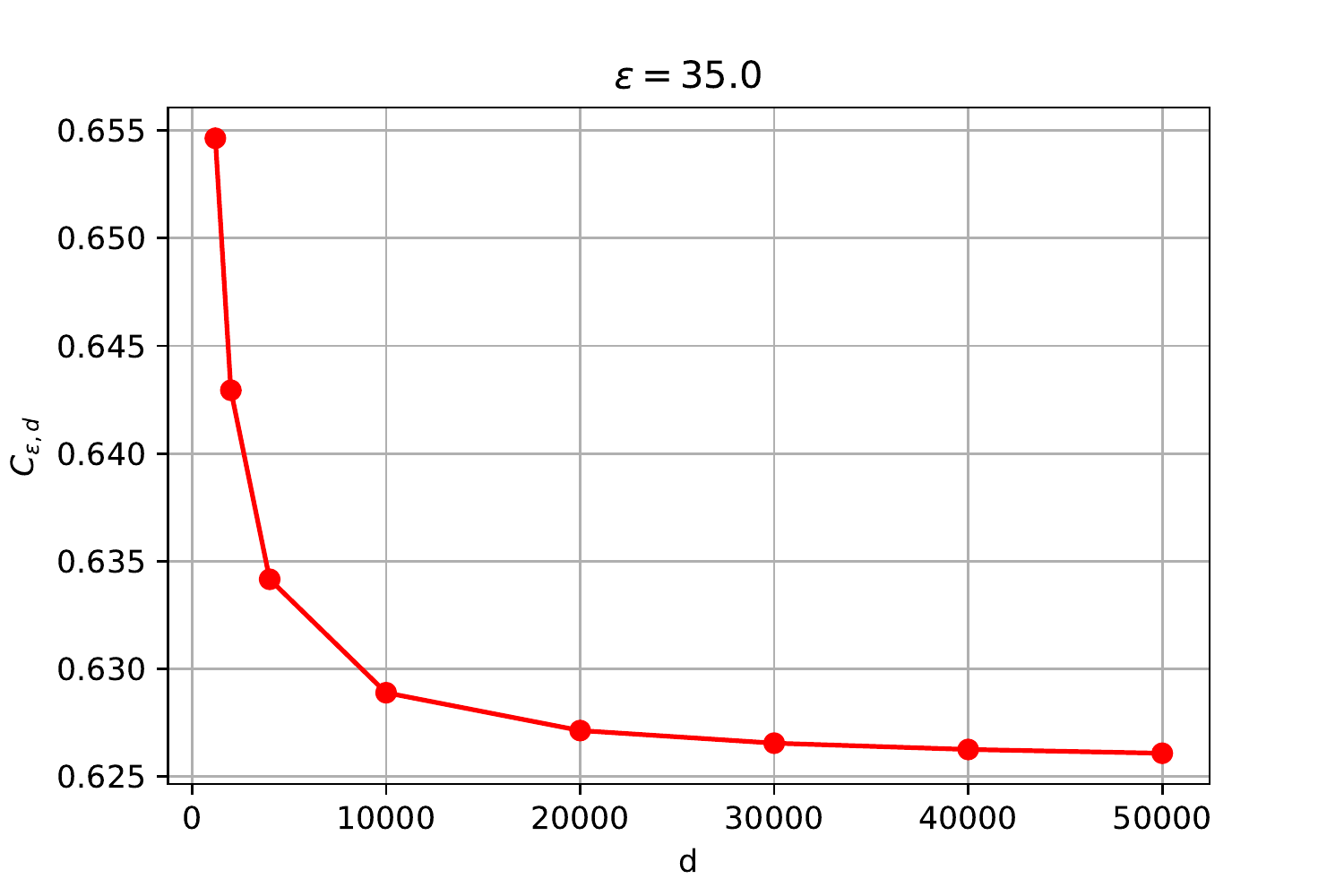}
      \end{overpic} &
      \hspace{-1cm}
      \begin{overpic}[width=.45\columnwidth]{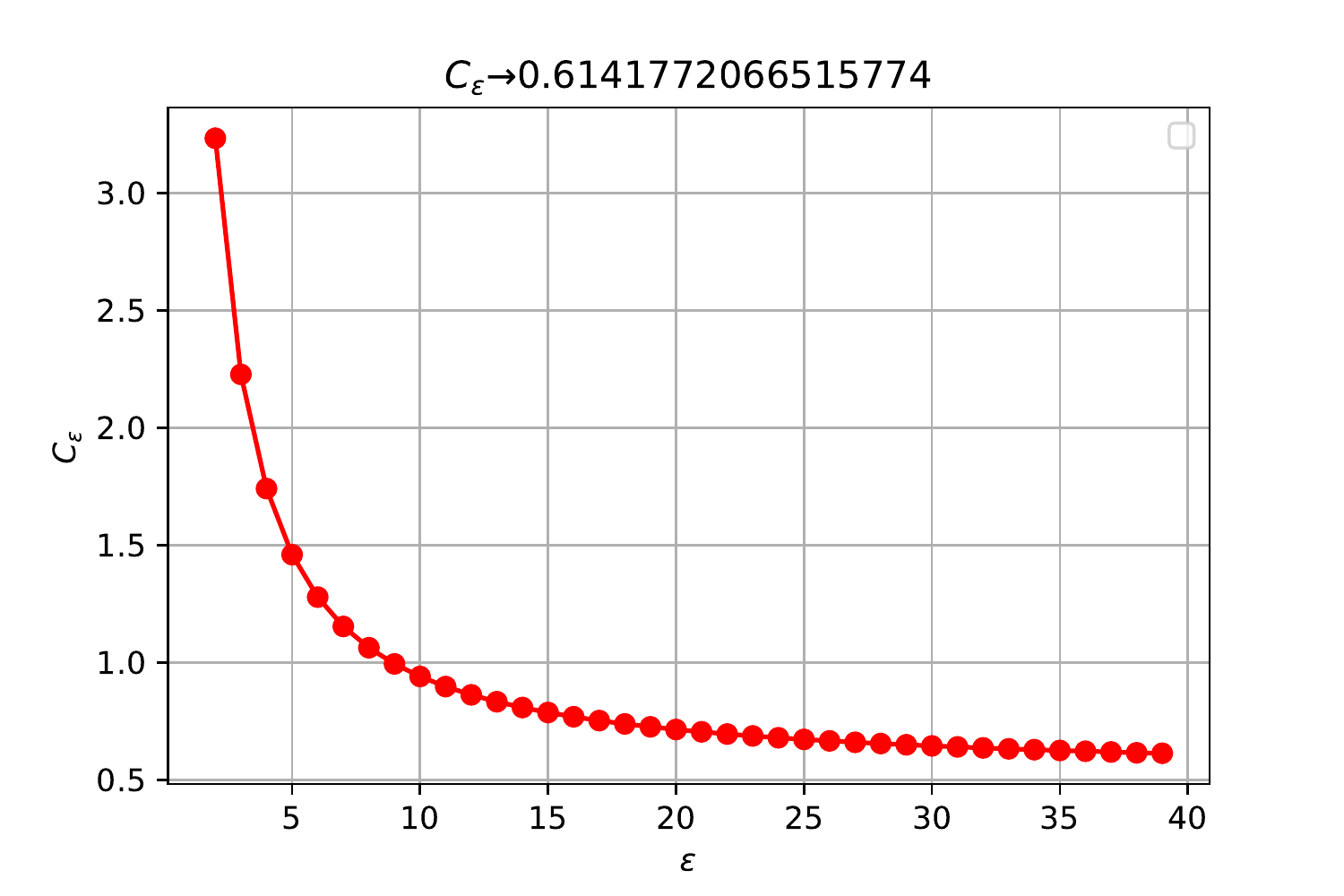}
      \end{overpic} \\
      (a) & (b) 
    \end{tabular}
    \caption{\label{fig:C-eps} (a) $C_{\diffp,d}$ as a function of $d$ for $\diffp = 35$. (b) $C_\diffp$ as a function of $\diffp$ (we approximate $C_\diffp$ by taking a sufficiently large dimension $d = 5 \cdot 10^4$)}
  \end{center}

\end{figure}
\fi

The following proposition shows that $C_{\diffp,d}$ converges as we increase the dimension $d$. We provide the proof in~\Cref{sec:proof-prop-C_eps}.
\begin{proposition}
\label{prop:C_eps}
    Fix $\diffp > 0$.
    For any $1 \le d_1 \le d_2$,
    \begin{equation*}
        1 - O \left(\frac{\diffp + \log d_2}{d_2} + \frac{\diffp}{d_1}\right) \le
        \frac{C_{\diffp,d_1}}{C_{\diffp,d_2}}
        \le 1 + O \left( \frac{\diffp + \log d_1}{d_1} + \frac{\diffp}{d_2}  \right).
    \end{equation*}
    \ificml
    In particular, $C_{\diffp,d} \stackrel{d \to \infty}{\to} C_\diffp$.
    \else
    In particular, as $d \to \infty$
    \begin{equation*}
        C_{\diffp,d} \stackrel{d \to \infty}{\to} C_\diffp.
    \end{equation*}
    \fi
\end{proposition}

The following proposition shows that $C_\diffp$ also converges as we increase $\diffp$. We present the proof in~\Cref{sec:proof-C_eps_limit}.
\begin{proposition}
\label{prop:C_eps_limit}
    There is $C^\star > 0$ such that $\lim_{\diffp \to \infty} C_\diffp = C^\star$.
\end{proposition}

\printbibliography

\appendix

\section{Missing details for $\pu$ (\Cref{sec:opt})}
\label{sec:apdx-pu}

\ificml
\subsection{Full details for $\pu$}

\begin{algorithm}
	\caption{$\pu(p,\gamma)$}
	\label{alg:pu2}
	\begin{algorithmic}[1]
		\REQUIRE $v \in \sphere^{d-1}$, $\gamma \in [0,1]$, $p \in [0,1]$. $B(\cdot; \cdot, \cdot)$ below is the incomplete Beta function $B(x;a,b) = \int_0^x t^{a-1}(1-t)^{b-1} \textrm{d}t$ and $B(a,b) = B(1; a, b)$.
		\STATE Draw $z \sim \mathsf{Ber}(p)$
		\IF{$z=1$}
		    \STATE Draw $V \sim \uniform \{u \in \sphere^{d-1}: \< u,v\> \ge \gamma \}$
		\ELSE
		     \STATE Draw $V \sim \uniform \{u \in \sphere^{d-1}: \< u,v\> < \gamma \}$
		\ENDIF
		\STATE Set $\alpha = \frac{d-1}{2}$ and $\tau = \frac{1 + \gamma}{2}$
		\STATE Calculate normalization constant
		\begin{equation*}
		    m = \frac{(1-\gamma^2)^\alpha}{2^{d-2} (d-1)} \left( \frac{p}{B(\alpha,\alpha) - B(\tau; \alpha,\alpha)} + \frac{1-p}{B(\tau; \alpha,\alpha)}   \right)
		\end{equation*}
        \STATE Return $\frac{1}{m} \cdot V$
	\end{algorithmic}
\end{algorithm}

\fi

\ificml
\subsection{Proof of~\Cref{lemma:symmetry}}
Given $\R$, we define $\R'$ as follows. First, sample a random rotation matrix $U \in \r^{d \times d}$ where $U^T U = I$, then set
\begin{equation*}
        \R'(x) = U^T \R(Ux).
\end{equation*}

We now prove that $\R'$ satisfies the desired properties. First, note that the privacy of $\R$ immediately implies the same privacy bound for $\R'$.
Moreover, we have that $\E[\R'(v)] = \E[U^T \E[\R(Uv)]] = \E[U^T Uv] = v$ as $\R$ is unbiased and $U^T U =I$. For utility, note that
\begin{align*}
    \err(\R')
        & = \sup_{v \in \sphere_2^{d-1}} \E_{\R'} \left[ \ltwo{\R'(v) - v}^2 \right] \\  
        & = \sup_{v \in \sphere_2^{d-1}} \E_{U, \R} \left[ \ltwo{U^T \R(Uv) - U^T Uv}^2 \right] \\
        & = \sup_{v \in \sphere_2^{d-1}} \E_{U, \R} \left[ \ltwo{\R(Uv) - Uv}^2 \right] \\
        & = \sup_{v \in \sphere_2^{d-1}, U} \E_{\R} \left[ \ltwo{\R(Uv) - Uv}^2 \right] \\
        & \le \sup_{v \in \sphere_2^{d-1}} \E_{\R} \left[ \ltwo{\R(v) - v}^2 \right] \\
        & = \err(\R).
\end{align*}
    
Finally, we show that the distributions of $\R'(v_1)$ and $\R'(v_2)$ are the same up to rotations. Indeed, let $V \in \r^{d \times d}$ be a rotation matrix such that $v_1 = Vv_2$. We have that $\R'(v_2)= U^T \R(Uv_2)$, which can also be written as
$\R'(v_2) \eqd (UV)^T \R(UVv_2) \eqd V^T U^T \R(Uv_1) \eqd V^T \R'(v_1)$ as $UV$ is also a random rotation matrix.
    
Now we prove the final property. Assume towards a contradiction there is $v_1 \in \sphere_2^{d-1}$ and $u_1,u_2 \in \r^d$ with $\ltwo{u_1} = \ltwo{u_2}$ such that $\frac{\pdf_{\R'(v_1)}(u_1)}{\pdf_{\R'(v_1)}(u_2)} > e^\diffp$. We will show that this implies that $\R'$ is not $\diffp$-DP. In the proof above we showed that $\R'(v_2) \eqd V^T \R'(v_1)$ for $v_1 = Vv_2$. Therefore for $V$ such that $V^T u_1 = u_2$ we get that $\pdf_{\R'(v_2)}(u_2) = \pdf_{\R'(v_1)}(u_1)$ which implies that 
\begin{equation*}
        \frac{\pdf_{\R'(v_2)}(u_2)}{\pdf_{\R'(v_1)}(u_2)} > e^\diffp,
\end{equation*}
which is a contradiction.
\fi

\subsection{Proof of~\Cref{lemma:two-prob} (missing details)}
\label{sec:apdx-two-prop-proof}
\newcommand{\Bbad}{B_{\mathsf{bad}}}
\newcommand{\pbad}{p_{\mathsf{bad}}}
\newcommand{\Vbad}{V_{\mathsf{bad}}}
\newcommand{\Va}{V_{A_0}}
Here we complete the missing details from the proof of~\Cref{lemma:two-prob}. We have $B_1,\dots,B_K$ sets such that for at least $K-d-2$ of the sets $B_i$ we have $\pdf(u_i) \in \{e^{\diffp},e^{-\diffp}\}  p$. We now show how to manipulate the probabilities on the other sets to satisfy our desired properties while not affecting the accuracy. 
Assume without loss of generality that $B_1,\dots,B_{d-2}$ do not satisfy this condition and let $\Bbad = \cup_{1 \le i \le d-2} B_i$. We now show that we can manipulate the density on $\Bbad$ such that it will satisfy this condition. Moreover, we show that the probability of $u \in \Bbad$ is small so that this does not affect the accuracy of the algorithm.
      Note that the probability $ u \in \Bbad$ is at most
      \begin{equation*}
      \pbad
          \defeq P(\hat \R \in \Bbad)
          \le d \max_{i} V_i p e^\diffp  
          \le d V \delta^d p e^\diffp,
      \end{equation*}
      where $V = \int_{u \in \sphere^{d-1}} 1 du$. Now we proceed to balance the density in $\Bbad$. Let $\Bbad = A_0 \cup A_1$ where $A_0 = A_1^-$ and $A_0 \cap A_1 = \emptyset$. We show how to balance the probabilities for $u \in A_0$ such that the mass on $A_0$ is $\pbad/2$. Then we define the density for $u \in A_1$ using the density of $u^-$ as $u^- \in A_0$.
      Let $\Va = \int_{u \in A_0} 1 du$ be the measure of the set $A_0$. We divide $A_0$ to two sets $A_0^1$ and $A_0^2$ such that $\int_{u \in A_0^1} 1 du = \rho \Va$ and $\int_{u \in A_0^2} 1du = (1-\rho) \Va$. We define a new density function $\tilde \pdf$ such that
      \begin{equation*}
        \tilde \pdf(u) = 
        \begin{cases}
                \hat \pdf(u) & \quad \quad \quad \text{if  } u \notin \Bbad \\
                \hat pe^\diffp & \quad \quad \quad \text{if  } u \in A_0^1 \\
                \hat p e^{-\diffp} & \quad \quad \quad \text{if  } u \in A_0^2 \\
                \tilde \pdf(u^-) & \quad \quad \quad \text{if  } u \in A_1
        \end{cases}
      \end{equation*}
      First, note that by design we have $\tilde f(u) = \tilde f(u^-)$.
      We now prove that the choice $\rho = \frac{{\pbad}/{2p\Va} - e^{-\diffp}}{e^{\diffp} - e^{-\diffp}}$ satisfies all of our conditions.
      First we show that $\rho \in [0,1]$. Indeed as $\hat f(u) \in [e^{-\diffp},e^\diffp] p$ for all $u$, we get that the average density in $\Bbad$ is also in this range, that is, $\frac{\pbad}{p\Vbad} = \frac{\pbad}{2p\Va} \in [e^{-\diffp},e^\diffp]$, which implies $\rho \in [0,1]$. Moreover, note that
      \begin{align*}
      \int_{u \in A_0} \tilde f(u) du 
             = pe^\diffp \Va \rho + pe^{-\diffp} \Va (1-\rho) 
             = \pbad/2.
      \end{align*}
      This implies that $\int_{u \in \sphere^{d-1}} \tilde f(u) du = 1$. Finally, note that this does not affect $\alpha$ too much as we have
      \begin{align*}
      \tilde \alpha 
        & = \sum_{u \in S} \tilde \pdf(u) V_i \<v, \bar u_i\> \\
        & = \sum_{u \in S \setminus \Bbad} \hat \pdf(u) V_i \<v, \bar u_i\> + \sum_{u \in S \cap \Bbad} \tilde \pdf(u) V_i \<v, \bar u_i\> \\
        & = \sum_{u \in S} \hat \pdf(u) V_i \<v, \bar u_i\>  + \sum_{u \in S \cap  \Bbad} \tilde \pdf(u) V_i \<v, \bar u_i\>  -  \sum_{u \in S \cap \Bbad} \hat \pdf(u) V_i \<v, \bar u_i\>\\
        & \ge \hat \alpha - \pbad.
      \end{align*}
      Note that for sufficiently small $\delta$, the error of this algorithm is now
      \begin{align*}
      \frac{1}{\tilde \alpha^2}
          & \le \frac{1}{(\hat \alpha - \pbad)^2} \\
          & \le \frac{1}{(\alpha - \pbad - 2 \delta)^2} \\
          & \le \frac{1}{\alpha^2} + \frac{20(\pbad + 2 \delta)}{\alpha^2} \\
          & \le \frac{1}{\alpha^2} + \tau,
      \end{align*}
      where the last inequality follows by choosing $\delta$ small enough such that $ \frac{20(\pbad + 2 \delta)}{\alpha^2} \le \tau$ which gives the claim.

\section{Proofs and missing details for~\Cref{sec:pug}}

\ificml
\subsection{Full details for $\pug$}

\begin{algorithm}
	\caption{$\pug(p,q)$}
	\label{alg:puG}
	\begin{algorithmic}[1]
		\REQUIRE $v \in \sphere^{d-1}$, $q \in [0,1]$, $p \in [0,1]$
		\STATE Draw $z \sim \mathsf{Ber}(p)$
		\STATE Let $U = \normal(0,\sigma^2) $ where $\sigma^2 = 1/d$
		\STATE Set $\gamma = \Phi_{\sigma^2}^{-1}(q) = \sigma \cdot \Phi^{-1}(q)$
		\IF{$z=1$}
		    \STATE Draw $\alpha \sim U \mid U \ge \gamma$
		\ELSE
		     \STATE Draw $\alpha \sim U \mid U < \gamma$
		\ENDIF
		\STATE Draw $V^{\perp} \sim \normal(0, \sigma^2(I - v v^T)$
		\STATE Set $V = \alpha v + V^{\perp}$
		\STATE Calculate
		\begin{equation*}
		    m = \sigma \phi(\gamma/\sigma) \left(\frac{p}{1 - q}  - \frac{1-p}{q} \right)
		\end{equation*}
        \STATE Return $\frac{1}{m} \cdot V$
	\end{algorithmic}
\end{algorithm}
\fi

\subsection{Proof of~\Cref{prop:pug-guarantees}}
\label{sec:proof-prop-pug-guarantees}
We will use the following helper lemma.
\begin{lemma}
\label{lemma:gaus-cond-exp}
    Let $U \sim \normal(0,\sigma^2)$. Then 
    \begin{equation*}
        \E[U \indic{ U \ge \gamma}] = \sigma \phi(\gamma/\sigma).
    \end{equation*}
\end{lemma}
\begin{proof}
    We have
    \begin{align*}
    \E[U \indic{ U \ge \gamma}]
        & = \int_{t=\gamma}^\infty u \phi_{\sigma^2}(t) dt \\
        & = \int_{t=\gamma}^\infty u \frac{1}{\sqrt{2 \pi \sigma^2}} e^{-t^2/2\sigma^2} dt \\
        & = \frac{\sigma}{\sqrt{2 \pi}} e^{-\gamma^2/2\sigma^2} \\
        & = \sigma \phi(\gamma/\sigma).
    \end{align*}
\end{proof}

We begin by proving $\pug$ is unbiased. Note that $\E[\frac{1}{m}V] = \E[\alpha/m] \cdot v$ therefore we need to show that $\E[\alpha]=m$. To this end, 
\begin{align*}
    \E[\alpha]
        & = \left( p \E[U \mid U \ge \gamma] + (1-p) \E[U \mid U < \gamma] \right)  \\
        & = \frac{p}{P(U \ge \gamma)} \E[U \indic{ U \ge \gamma}] + \frac{1-p}{P(U < \gamma)} \E[U \indic{ U \le \gamma}] \\
        & =   \E[U \indic{ U \ge \gamma}] \left( \frac{p}{P(U \ge \gamma)}  - \frac{1-p}{P(U < \gamma)} \right) \\
        & = \sigma \phi(\gamma/\sigma) \left( \frac{p}{P(U \ge \gamma)}  - \frac{1-p}{P(U < \gamma)} \right) \\
        & = m,
\end{align*}
where the third inequality follows since $\E[U] = \E[U \indic{ U \ge \gamma}] + \E[U \indic{ U < \gamma}] = 0$, and the last inequality follows since~\Cref{lemma:gaus-cond-exp} gives that $\E[U \indic{ U \ge \gamma}] = \sigma \phi(\gamma/\sigma)$ for $U \sim \normal(0,\sigma^2)$. For the claim about utility, as $\pug$ is unbiased we have
\begin{equation*}
    \E[\ltwo{\pug_2(v) - v}^2] 
        =  \E[\ltwo{\pug_2(v)}^2] - 1
        = \frac{1}{m^2} \left( \E [\alpha^2] + \frac{d-1}{d} \right) - 1
\end{equation*}
Now we prove the claim about the limit. First, note that $P(U \le \gamma) = q$ hence we can write 
\begin{equation*}
    m = \sigma \phi(\gamma/\sigma) \left( \frac{p}{1-q}  - \frac{1-p}{q} \right). 
\end{equation*}
Moreover, \Cref{lemma:bound-sec-moment} and~\Cref{lemma:ub-gamma} shows that $\E[\alpha^2] \le O \left(\gamma^2 + \frac{\diffp + \log d}{d} \right) \le O \left(\frac{\diffp + \log d}{d} \right) $. Taking limit as $d \to \infty$, this yields that
\begin{equation*}
     m^2 \cdot \E[\ltwo{\pug_2(v) - v}^2] 
        \stackrel{d \to \infty}{\to} 1.
\end{equation*}

Now we proceed to prove the privacy claim. We need to show that for every $v_1,v_2 \in \sphere_2^d$ and $u \in \r^d$
\begin{equation*}
        \frac{\pdf_{\pug(v_1)}(u)}{\pdf_{\pug(v_2)}(u)} \le e^\diffp.
\end{equation*}
For every input vector, we divide the output space to two sets: $S_v = \{u \in \r^d : \<u,v\> \ge \gamma \}$ and $S_v^c = \r^d \setminus S_v$. The definition of $\pug$ implies that for $u \in S_v$ we have 
\begin{equation*}
    \pdf_{\pug(v)}(u) = p \frac{\pdf_U(u)}{P(U \ge \gamma)},
\end{equation*}
and for $u \notin S_v$ we have
\begin{equation*}
    \pdf_{\pug(v)}(u) = (1-p) \frac{\pdf_U(u)}{P(U \le \gamma)}.
\end{equation*}
Using the notation $q = P(U \le \gamma)$, we now have that for any $v_1,v_2$ and $u$
\begin{align*}
    \frac{\pdf_{\pug(v_1)}(u)}{\pdf_{\pug(v_2)}(u)} 
        & \le \frac{p/(1-q)}{(1-p)/q} \\
        & \le \frac{p}{1-p} \cdot \frac{q}{1-q} \\
        & \le e^\diffp,
\end{align*}
where the first inequality follows since we must have $u \in S_{v_1}$ and $u \notin s_{v_2}$ to maximize the ratio.


\subsection{Helper lemmas for~\Cref{thm:pug-opt}}

\begin{lemma}
\label{lemma:bound-diff-exp}
    We have
    \begin{equation*}
    |\E[W \indic{W \ge \gamma }] - \E[U \indic{U \ge \gamma }]|     
        \le O\left(\frac{\sqrt{\log d}}{d^{3/2}}\right).
    \end{equation*}
\end{lemma}
\begin{proof}
    \ificml
    We use the fact that for $U$ and $W$, we have $P(U \le \gamma) \le  P(W \le \gamma) + \frac{8}{d-4}$~(\citealp[Inequality (1)]{DiaconisFr87}). 
    \else
    We use the fact that for $U$ and $W$, we have $P(U \le \gamma) \le  P(W \le \gamma) + \frac{8}{d-4}$~(\cite[Inequality (1)]{DiaconisFr87}). 
    \fi
    We have
    \begin{align*}
    |\E[W \indic{W \ge \gamma }] - \E[U \indic{U \ge \gamma }]|     & = |\int_{\gamma}^\infty u (\pdf_W(u) - \pdf_U(u)) du| \\
       & \le | \int_{\gamma}^C u (\pdf_W(u) - \pdf_U(u)) du|  + | \int_{C}^\infty u (\pdf_W(u) - \pdf_U(u)) du| \\
       & \le C \norm{P_W - P_U}_{TV} + \int_{C}^1 u \pdf_W(u) du + \frac{\pdf_U(C)}{d} \\
       & \le \frac{8C}{d-4} + \frac{e^{-C^2 d/2}}{\sqrt{2 \pi d}} + \pdf_W(C) \\
       & \stackrel{(i)}{\le} \frac{8C}{d-4} + \frac{e^{-C^2 d/2}}{\sqrt{2 \pi d}} + O(\sqrt{d} e^{-C^2 d/8}) \\
       & \le O(\frac{\sqrt{\log d}}{d^{3/2}}).
    \end{align*}
    where the last inequality follows by setting $C = \Theta(\sqrt{\frac{\log(d)}{d}})$.
    Inequality $(i)$ follows since $W$ has the same distribution as $2B-1$ for $B \sim \mathsf{Beta}(\frac{d-1}{2},\frac{d-1}{2})$. Indeed setting $\alpha = \frac{d-1}{2}$, we have for any $b = 1/2 + \rho$
    \begin{align*}
    \pdf_B(b)
         & = \frac{(b(1-b))^{\alpha-1}}{B(\alpha,\alpha)} \\
         & = \frac{(1-\rho^2)^{\alpha-1}}{4^{\alpha-1}} \frac{1}{B(\alpha,\alpha)} \\
         & \le \frac{e^{-\rho^2(\alpha-1)}}{4^{\alpha-1}} \frac{1}{B(\alpha,\alpha)} \\
         & \le O(\sqrt{\alpha} e^{-\rho^2\alpha}),
    \end{align*}
    where the last inequality follows from
    \begin{align*}
    B(\alpha,\alpha)
         = \frac{2}{\alpha} \frac{1}{\binom{2\alpha}{\alpha}} 
        \ge \frac{2}{\alpha} \Omega(\frac{\sqrt{\alpha}}{4^\alpha}) 
        = \Omega(\frac{1}{\sqrt{\alpha} 4^\alpha}) .
    \end{align*}
    The claim now follows as $\pdf_W(C) = \pdf_B(1/2 + C/2)$.
\end{proof}


\begin{lemma}
\label{lemma:bound-sec-moment}
    We have
    \begin{equation*}
        \E[\alpha^2] \le O \left(\gamma^2 + \frac{\diffp_G + \log d}{d} \right).
    \end{equation*}
\end{lemma}
\begin{proof}
    It is enough to upper bound $\E[U^2 \mid U \ge \gamma]$. Since $P(U \ge \gamma) \ge e^{-\diffp_G}$
    \begin{align*}
    \E[U^2 \mid U \ge \gamma]
        & = \int_{\gamma}^C u^2 \frac{f_U(u)}{P(U \ge \gamma)} du + \int_{C}^\infty u^2 \frac{f_U(u)}{P(U \ge \gamma)} du \\
        & \le C^2 + e^{\diffp_G} \frac{1}{\sqrt{2\pi d}} \int_{C}^\infty u^2 de^{-u^2 d/2} du \\
        & \le  C^2  + e^{\diffp_G} \frac{e^{-C^2 d/4}}{\sqrt{2\pi d}} \int_{C}^\infty u^2 de^{-u^2 d/4} du \\ 
        & \le O \left(\gamma^2 + \frac{\diffp_G + \log d}{d} \right) ,
    \end{align*}
    where the last inequality follows by setting $C = \Theta\left( \max \left(\gamma, \sqrt{\frac{\diffp_G + \log d}{d}} \right) \right)$.
\end{proof}

The following lemma is also useful for our analysis.
\begin{lemma}
\label{lemma:ub-gamma}
    Assume $\pug(p,\gamma)$ is $\diffp$-DP. Then $\gamma^2 \le \frac{2 \log(e^\diffp + 1)}{d}$.
\end{lemma}
\begin{proof}
    \Cref{prop:pug-guarantees} implies that $\gamma = \frac{\Phi^{-1}(q)}{\sqrt{d}}$ where $q = \frac{e^{\diffp_1}}{e^{\diffp_1} + 1}$ for $\diffp_1 \le \diffp$. As $\Phi^{-1}(q)$ is increasing in $q$, we have that 
    \begin{equation*}
    \gamma 
        = \frac{\Phi^{-1}(q)}{\sqrt{d}} 
        = \frac{\Phi^{-1}(\frac{e^{\diffp}}{e^{\diffp} + 1})}{\sqrt{d}} .
    \end{equation*}
    Gaussian concentration gives that for $t^2 \ge 2 \log(e^\diffp + 1)$
    \begin{equation*}
     P(\normal(0,1) > t)
        \le e^{-t^2/2} 
        \le 1/(e^\diffp + 1).
    \end{equation*}
    This implies that $\Phi^{-1}(\frac{e^{\diffp}}{e^{\diffp} + 1}) \le \sqrt{2 \log(e^\diffp + 1)} $ and proves the claim.
    
\end{proof}

\ificml
\subsection{Proof of~\Cref{thm:pug-opt}}
\label{sec:proof-pug-opt}
    The privacy proof is straightforward as both algorithms use the same $p$ and $q$ therefore enjoy the same privacy parameter. Moreover, the second part of the claim follows immediately from the first part. 
    
    Now we prove the first part of the claim. Let $\gamma_2$ be such that $q = P(W \le \gamma_2)$ where $W \sim \uniform(\sphere_2^{d-1})$. Then we have
    \begin{align*}
    m_2
        & = \E[W \indic{W \ge \gamma_2 }] \frac{p+q-1}{q(1-q)}  \\
    \err_2 & = \err(\pu(p,q))  = \frac{1}{m_2^2}    .
    \end{align*}
    Similarly for $\pug$ we have (see e.g. proof of~\Cref{prop:pug-guarantees})
    \begin{align*}
    m_G
         & = E[\alpha]
           = \E[U \indic{U \ge \gamma_G }] \frac{p+q-1}{q(1-q)} \\
    \err_G & = \err(\pug(p,q)) = \frac{\E[\alpha^2]-1/d}{m_G^2} + \frac{1}{m_G^2}  .
    \end{align*}
    Therefore we have
    \begin{align}
    \label{eq:err-ratio}
    \sqrt{\frac{\err_G}{\err_2}}
        & \le \frac{m_2}{m_G} \sqrt{1 + \E[\alpha^2]} \nonumber \\
        & \le \frac{\E[W \indic{W \ge \gamma_2 }]}{\E[U \indic{U \ge \gamma_G }]} \sqrt{1 + \E[\alpha^2]} \nonumber \\
        & \stackrel{(i)}{\le}  \frac{\E[W \indic{W \ge \gamma_2 }]}{\E[U \indic{U \ge \gamma_G }]}  \left(1 +  O\left( \gamma_G + \sqrt{\frac{\diffp + \log d}{d}}\right) \right),
    \end{align}
    where inequality $(i)$ follows~\Cref{lemma:bound-sec-moment}. 
    We now upper bound the first term. Note first that for the same $\gamma$ we have
    \begin{align*}
     \frac{\E[W \indic{W \ge \gamma }]}{\E[U \indic{U \ge \gamma}]} 
        & \stackrel{(i)}{\le} 1 + O \left( \frac{ {\sqrt{\log d}}{}}{d^{3/2} \E[U \indic{U \ge \gamma }]} \right) \\
        & \stackrel{(ii)}{\le} 1 + O \left( \frac{ \sqrt{\log d}}{d} \right),
    \end{align*}
    where $(i)$ follows from~\Cref{lemma:bound-diff-exp} and $(ii)$ follows since $\E[U \indic{U \ge \gamma }] \ge \E[U \indic{U \ge 0}] = \E[|U|]/2 \ge \Omega(\sigma) = \Omega(1/\sqrt{d}) $.
    We will need one more step to finish the proof as $\gamma_2$ and $\gamma_G$ are not necessarily equal. 
    We divide to cases. If $\gamma_2 \ge \gamma_G$, then $\E[U \indic{U \ge \gamma_G }] \ge \E[U \indic{U \ge \gamma_2 }]$ and therefore
    \begin{align*}
     \frac{\E[W \indic{W \ge \gamma_2 }]}{\E[U \indic{U \ge \gamma_G }]}
        & \le \frac{\E[W \indic{W \ge \gamma_2 }]}{\E[U \indic{U \ge \gamma_2 }]}.
    \end{align*}
    Similarly, if $\gamma_2 \le \gamma_G$  then $\E[W \indic{W \ge \gamma_2 }] \le \E[W \indic{W \ge \gamma_G }]$ and therefore
    \begin{align*}
     \frac{\E[W \indic{W \ge \gamma_2 }]}{\E[U \indic{U \ge \gamma_G }]}
        & \le \frac{\E[W \indic{W \ge \gamma_G }]}{\E[U \indic{U \ge \gamma_G }]}.
    \end{align*}
    Overall this proves that 
    \begin{align*}
     \frac{\E[W \indic{W \ge \gamma_2 }]}{\E[U \indic{U \ge \gamma_G }]}
        & \le 1 + O \left( \frac{ \sqrt{\log d}}{d} \right).
    \end{align*}
    Putting this back in inequality~\eqref{eq:err-ratio}, we have
    \begin{align*}
    \sqrt{\frac{\err_G}{\err_2}}
        & \le \left( 1 + O \left( \frac{ \sqrt{\log d}}{d} \right) \right)\left(1 +  O\left( \gamma_G + \sqrt{\frac{\diffp + \log d}{d}}\right) \right) \\
        & \le 1 + O \left( \sqrt{\frac{\diffp + \log d}{d}} \right),
    \end{align*}
    where the last inequality follows from~\Cref{lemma:ub-gamma} as $\gamma \le O(\sqrt{{(\diffp+1)}/{d}})$. The claim follows.
\fi

\subsection{Proof of~\Cref{prop:C_eps}}
\label{sec:proof-prop-C_eps}
Recall that the error of $\pug(p,q)$ for dimension $d$ is 
\begin{equation*}
        \err_d(\pug(p,q)) = \frac{\E[\alpha^2] + 1}{m_d^2} - 1    
\end{equation*}
where 
\begin{equation*}
	    m_d = \frac{ \phi(\gamma\sqrt{d})}{\sqrt{d}} \left(\frac{p}{1-q}  - \frac{1-p}{q} \right).
\end{equation*}
Note first that $\gamma \sqrt{d} =  \Phi^{-1}(q)$ which immediately implies that
\begin{equation*}
	    \frac{m_{d_1}}{m_{d_2}} = \sqrt{\frac{d_2}{d_1}}.
\end{equation*}
\Cref{lemma:bound-sec-moment} and~\Cref{lemma:ub-gamma} show that $\E[\alpha^2] \le O ( \frac{\diffp + \log d}{d})$. Finally, as $\err_d(p,q) \ge C d/\diffp$ for all $p$ and $q$, this implies that $m_d^2 \le O(\diffp/d)$.
Letting $\err_d$ denote the error for for inputs $u\in \sphere^{d-1}$, we now have
\begin{align*}
    \frac{C_{\diffp,d_1}}{C_{\diffp,d_2}}
        & = \frac{\err_{d_1}(\pug(p_1,q_1))/d_1}{\err_{d_2}(\pug(p_2,q_2))/d_2} \\
        & \le \frac{d_2}{d_1} \frac{\err_{d_1}(\pug(p_2,q_2))}{\err_{d_2}(\pug(p_2,q_2))} \\
        & \le \frac{d_2}{d_1} \frac{ \frac{\E[\alpha_1^2] + 1}{m_{d_1}^2} } {\frac{ 1}{m_{d_2}^2} - 1 } \\
        & \le \frac{d_2}{d_1} (\E[\alpha_1^2] + 1) \frac{m_{d_2}^2} {m_{d_1}^2} \frac{1}{1 - m_{d_2}^2 } \\
        & = (\E[\alpha_1^2] + 1) \frac{1}{1 - m_{d_2}^2 } \\
        & \le \left( 1 + O( \frac{\diffp + \log d_1}{d_1}) \right)  \left( 1 + O( \frac{\diffp}{d_2}) \right) \\
        & \le 1 + O \left(  \frac{\diffp + \log d_1}{d_1} + \frac{\diffp}{d_2} \right).
\end{align*}
This proves the right hand side of the inequality. The left side follows using the same arguments by noting that 
\begin{align*}
    \frac{C_{\diffp,d_1}}{C_{\diffp,d_2}}
        & = \frac{\err_{d_1}(\pug(p_1,q_1))/d_1}{\err_{d_2}(\pug(p_2,q_2))/d_2} \\
        & \ge \frac{d_2}{d_1} \frac{\err_{d_1}(\pug(p_1,q_1))}{\err_{d_2}(\pug(p_1,q_1))}.
\end{align*}
The second part of the claim regarding the limit follows directly from the first part.

\subsection{Proof of~\Cref{prop:C_eps_limit}}
\label{sec:proof-C_eps_limit}
We need the following lemma for this proof.

\begin{lemma}
\label{lemma:c_eps-prop}
    We have
    \begin{enumerate}
        \item 
        $C_{\diffp_2} \le C_{\diffp_1} \frac{\diffp_2}{\diffp_1} $ for $\diffp_1 \le \diffp_2$
        \item $C_{k \diffp} \le C_\diffp$ for any integer $k \ge 1$
    \end{enumerate}
\end{lemma}

\begin{proof}
    The first part follows as $\err_{\diffp_2,d} \le \err_{\diffp_1,d}   $ as $\diffp_1 \le \diffp_2$ which implies that
    \begin{equation*}
     C_{\diffp_2,d} 
        = \diffp_2 \err_{\diffp_2,d}/d 
        \le \diffp_2 \err_{\diffp_1,d}/d 
        =  \frac{\diffp_2 }{\diffp_1} \diffp_1 \err_{\diffp_1,d}/d
        = \frac{\diffp_2 }{\diffp_1} C_{\diffp_1,d}.
    \end{equation*}
    This implies the claim for $C_\diffp$.
    
    The second part follows from apply a repetition-based randomizer. Given an $\diffp$-DP local randomizer $\R_d$ for $\sphere_2^{d-1}$ that achieves $C_{\diffp,d}$ (that is, $\err(\R_d) = C_{\diffp,d} d /\diffp$), the randomizer $\R'_d$ that returns an average of $k$ applications of $\R$ is $k \diffp$-DP and has error $ \err(R') \le \frac{\err(R)}{k}$. Thus we have
    \begin{equation*}
     C_{k\diffp,d} 
        = k \diffp \err_{k \diffp,d}/d 
        \le k \diffp \err(\R'_d)/d
        \le \diffp \err(\R_d)/d
        =  C_{\diffp,d}.
    \end{equation*}
    The proves the claim for $C_{\diffp}$ as it is true for all $d$.
\end{proof}

We are now ready to prove~\Cref{prop:C_eps_limit}.
\begin{proof}(\Cref{prop:C_eps_limit})
First, note that $C_\diffp$ is a sequence of real numbers that is bounded from below by zero. Thus $\liminf_{\diffp \to \infty} C_{\diffp}$ exists. 
Now we will show that $\lim_{\diffp \to \infty} C_\diffp$ exists. Let $C^\star = \liminf_{\diffp \to \infty} C_\diffp$. It is enough to show that for all $\delta>0$ there is $\diffp_0$ such that for all $\diffp \ge \diffp_0$ we get $C_\diffp \le C^\star + \delta$. The definition of $\liminf$ implies that there is $\diffp'$ such that $C_{\diffp'} \le C^\star + \delta/2$. We will show that for all $\diffp \ge k \diffp_1$ we get $C_\diffp \le C^\star + \delta$ for some large $k$. Let $\diffp = k' \diffp_1 + \diffp_2$ where $0 \le \diffp_2 < \diffp_1$ and $k' \ge k$. We have that
    \begin{align*}
    C_{\diffp} 
        & \stackrel{(i)}{\le} C_{k' \diffp_1} \frac{\diffp}{k'\diffp_1} \\
        & \stackrel{(ii)}{\le} C_{\diffp_1} \frac{\diffp}{k'\diffp_1} \\ 
        & \le C_{\diffp_1} (1 + 1/k) \\
        & \le (C^\star + \delta/2) (1 + 1/k) \\
        & \le C^\star + \delta,
    \end{align*}
    where $(i)$ and $(ii)$ follow from the first and second items of~\Cref{lemma:c_eps-prop} and the last inequality follows by setting $k = \ceil{1/(C^\star + \delta/2)}$. The claim follows.
    
\end{proof}



\end{document}